\documentclass[10pt,twocolumn,letterpaper]{article}

\usepackage[pagenumbers]{cvpr}
\usepackage{balance}
\usepackage{dsfont}

\usepackage{multibib}
\usepackage{etoolbox}
\makeatletter
\providecommand{\sf@counterlist}{}
\pretocmd{\multicols}{\let\sf@counterlist\@empty}{}{}
\makeatother
\newcites{app}{References}
\usepackage{graphicx}
\usepackage{svg}
\usepackage{booktabs}
\usepackage{rotating}
\usepackage{multirow} 
\usepackage{pifont}
\usepackage{textcomp}
\usepackage{stfloats}
\usepackage{url}
\usepackage{verbatim}
\usepackage{amsmath,amsfonts}
\usepackage{algorithmic}
\usepackage{array}
\usepackage[caption=false,font=normalsize,labelfont=sf,textfont=sf]{subfig}

\definecolor{cvprblue}{rgb}{0.21,0.49,0.74}
\usepackage[pagebackref,breaklinks,colorlinks,allcolors=cvprblue]{hyperref}
\def\paperID{13300} % *** Enter the Paper ID here
\def\confName{CVPR}
\def\confYear{2026}

\title{AnoRefiner: Anomaly-Aware Group-Wise Refinement for \\Zero-Shot Industrial Anomaly Detection}

\author{
Dayou Huang\textsuperscript{1} \quad
Feng Xue\textsuperscript{4} \quad
Xurui Li\textsuperscript{1} \quad
Yu Zhou\textsuperscript{1,2,3,$\dagger$} \\
$^1$School of Electronic Information and Communications,
Huazhong University of Science and Technology \\
$^2$ Hubei Key Laboratory of Smart Internet Technology,
Huazhong University of Science and Technology \\
$^3$Artificial Intelligence Research Institute, Wuhan JingCe Electronic Group Co., LTD \\
$^4$Department of Information Engineering and Computer Science, University of Trento
}

\begin{document}
\maketitle

\begin{abstract}
Zero-shot industrial anomaly detection (ZSAD) methods typically yield coarse anomaly maps as vision transformers (ViTs) extract patch-level features only.
To solve this, recent solutions attempt to predict finer anomalies using features from ZSAD,
but they still struggle to recover fine-grained anomalies without missed detections,
mainly due to the gap between randomly synthesized training anomalies and real ones.
We observe that anomaly score maps exactly provide complementary spatial cues that are largely absent from ZSAD's image features,
a fact overlooked before.
Inspired by this, we propose an anomaly-aware refiner (AnoRefiner) that can be plugged into most ZSAD models and improve patch-level anomaly maps to the pixel level.
First, we design an anomaly refinement decoder (ARD) that progressively enhances image features using anomaly score maps, reducing the reliance on synthetic anomaly data.
Second, motivated by the mass production paradigm,
we propose a progressive group-wise test-time training (PGT) strategy that trains ARD in each product group for the refinement process in the next group,
while staying compatible with any ZSAD method.
Experiments on the MVTec AD and VisA datasets show that AnoRefiner boosts various ZSAD models by up to a 5.2\% gain in pixel-AP metrics,
which can also be directly observed in many visualizations.
The code will be available at \href{https://github.com/HUST-SLOW/AnoRefiner}{https://github.com/HUST-SLOW/AnoRefiner}.
\end{abstract}
 
\section{Introduction}
\label{sec:intro}
\renewcommand{\thefootnote}{}
\footnotetext{$\dagger$ Corresponding Author.}
\renewcommand{\thefootnote}{\arabic{footnote}}
Industrial anomaly detection \cite{batzner2024efficientad,zavrtanik2021draem,zavrtanik2022dsr,zhang2023prototypical} is crucial for industrial applications such as quality control, automated visual inspection, and smart manufacturing, as it ensures product quality and operational efficiency.
However, the task remains challenging due to the diversity of products and the complexity of real-world anomalies, which vary significantly in appearance, size, and context.
These difficulties highlight the need for zero-shot anomaly detection (ZSAD) methods that can generalize without any labeled data.

\begin{figure}[t]
  \centering
  \setlength{\belowcaptionskip}{-7pt}
  \includegraphics[width=1\linewidth]{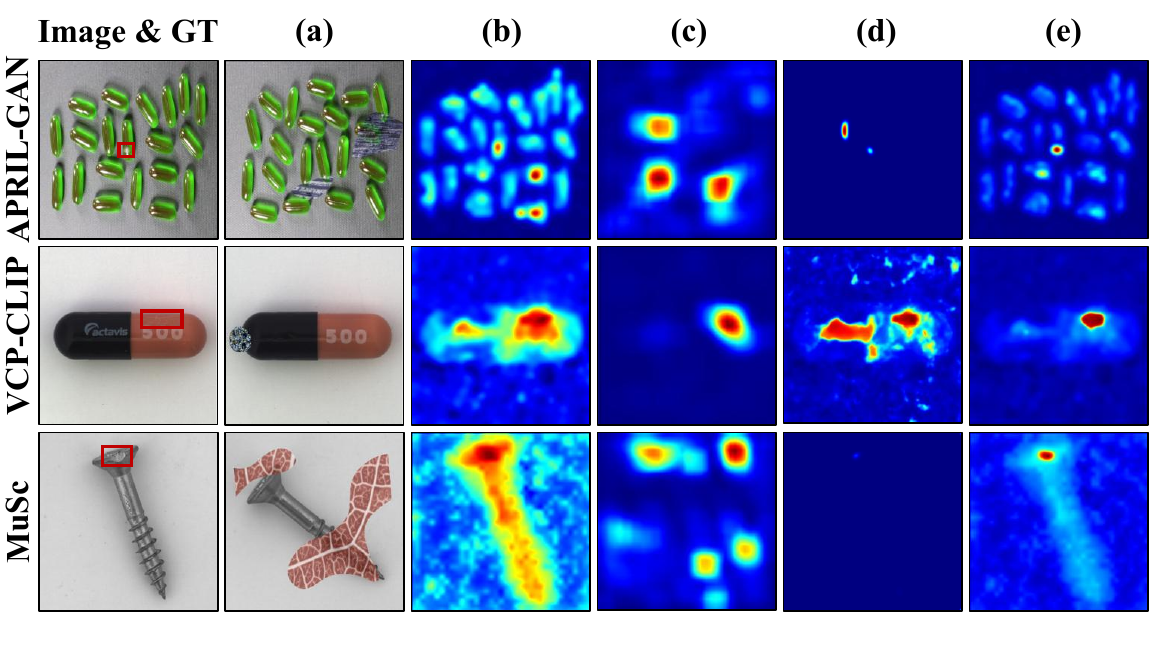}
  \caption{Comparison of anomaly maps generated by APRIL-GAN \cite{chen2023april}, VCP-CLIP \cite{qu2024vcp}, and MuSc \cite{li2024musc}. (a) Pseudo-anomaly images, (b) Coarse anomaly maps from ZSAD methods, (c) Refinement using DeSTseg \cite{zhang2023destseg} decoder, (d) Refinement using RealNet \cite{zhang2024realnet} decoder, and (e) Refinement result from our ARD.}
  \label{fig:introduction}
\end{figure}

Current ZSAD approaches could be broadly categorized into two paradigms.
One line of methods \cite{cao2025adaclip,chen2023april,deng2023anovl,qu2024vcp,zhou2023anomalyclip} leverages the generalization capability of pre-trained vision-language models like CLIP \cite{radford2021learning}, identifying anomalies by measuring cross-modal semantic similarities between text prompts and test images.
Another line \cite{li2024musc, li2024zero} focuses on inter-image analysis, detecting anomalies by comparing feature similarity among patches across different test images.
Nevertheless, due to the patch-wise feature extraction mechanism of ViTs \cite{dosovitskiy2020image},
these ZSAD methods only coarsely localize the anomalies,
which cannot be directly applied to the downstream applications like anomaly repair,
as shown in \cref{fig:introduction}(b).

A common and straightforward solution is to refine the localization of the anomalies by leveraging the refinement decoders \cite{miccai2015unet, zhang2023destseg, zhang2024realnet}. However, our experiments show that such a solution failed to produce satisfactory refinement results. 
As shown in 
\cref{fig:introduction}(c) and (d), several false positives and false negatives appear,
and we reveal the key implicit limitations as follows.
In industrial environments, real anomalous samples are typically too scarce to support effective supervised refinement network training, and thus existing approaches often leverage the generated random anomalies as the training samples, which may diverge significantly from real ones, as shown in \cref{fig:introduction}(a).
Such a discrepancy causes the learned refinement decoder to fail to generalize to real anomalies, 
and hence leads to a severe localization performance degradation.

\begin{figure}[t]
  \centering
  \setlength{\belowcaptionskip}{-7pt}
  \includegraphics[width=1\linewidth]{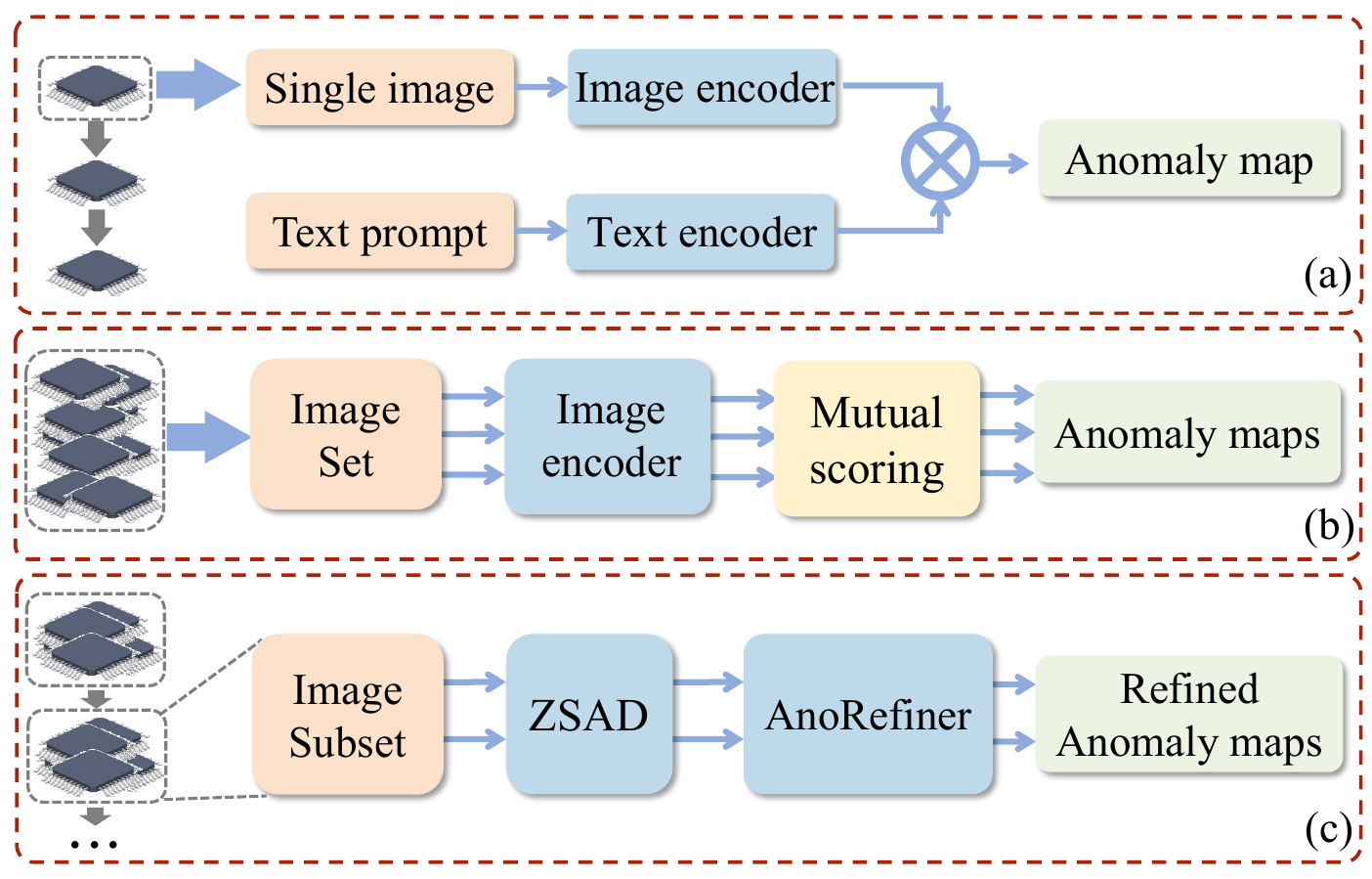}
  \caption{(a) CLIP-based methods operate on single test images. (b) Inter-image comparison methods leverage the entire test set for analysis. (c) Our AnoRefiner utilizes the current group to refine the anomaly localization in the next group.}
  \label{fig:introduction2}
\end{figure}

To address this issue, we propose a novel Anomaly-aware Refiner (AnoRefiner)  
to enhance the anomaly localization accuracy, 
which contains an anomaly-refinement decoder and a progressive group-wise test-time training strategy.
\textbf{Our design comes from the key observation that the anomaly score map is complementary to the decoded image features},
which is rarely discussed in existing approaches. 
Specifically, we propose an Anomaly Refinement Decoder (ARD), which progressively enhances the anomaly-aware decoded image features by using the anomaly score map, 
and further leverages both of them to cross-refine each other.
In addition, in mass production scenarios, industrial lines are typically designed to produce a large number of identical products, 
thus we propose a Progressive Group-wise Test-Time Training strategy (PGT) to train ARD, which is tailored to the current production paradigm.
Unlike single-image processing (\cref{fig:introduction2} (a)) or full-dataset processing (\cref{fig:introduction2} (b)), PGT trains ARD on the current group of product images at test time,
and refines the anomaly localization in the next group. 
In such a way, the proposed AnoRefiner acts as a plug-and-play module that can be integrated into most ZSAD frameworks, effectively transforming their patch-level anomaly maps into precise pixel-level localization.
This refinement process significantly suppresses both false positives and false negatives, as we show in \cref{fig:introduction}(e),
which indicates that AnoRefiner well alleviates the dependency on training samples.
For a comprehensive validation, we apply AnoRefiner to three different ZSAD approaches, 
and evaluate them on the MVTec AD dataset and VisA dataset.
The superior performance demonstrates the effectiveness of AnoRefiner,
it achieves pixel-AP improvement of 5.2\% and pixel-F1 improvement of 3.4\% on average.

Our main contributions can be summarized as follows:

\begin{itemize}
\item To the best of our knowledge, 
AnoRefiner is the first work that mitigates the reliance on high-quality labeled data, which is required by previous refinement decoders.

\item We propose two key components in our AnoRefiner: an Anomaly Refinement Decoder (ARD) that mutually refines anomaly maps and image features for precise localization, and a Progressive Group-wise Test-time Training (PGT) strategy that adapts the model using product-wise image groups.

\item Extensive experiments show that AnoRefiner generalizes across diverse ZSAD backbones, achieving average gains of +5.2\% in pixel-AP and +3.4\% in pixel-F1 on MVTec AD and VisA benchmarks.
\end{itemize}

\section{Related Work}
\label{sec:related work}

\subsection{Zero-shot Anomaly Detection}
Recent advances in foundational vision-language models like CLIP \cite{radford2021learning} have accelerated zero-shot anomaly detection (ZSAD) \cite{cao2023segment,ma2025aa,qu2024vcp}, which identifies anomalies without anomaly-specific training.
Current ZSAD methods generally follow two paradigms.
One line of work \cite{jeong2023winclip, deng2023anovl, cao2025adaclip, qu2025bayesian} leverages the generalization ability of the pre-trained CLIP model, identifying anomalies via cross-modal alignment between text prompts and test images.
Another line \cite{li2024musc, li2024zero} focuses on inter-image analysis, detecting anomalies by measuring patch-wise similarity across different test images.
A common limitation of the above methods is their dependence on low-resolution features from pre-trained Vision Transformers \cite{dosovitskiy2020image}, which often yield low-resolution anomaly maps and coarse segmentation.
To address this limitation, we propose the Anomaly-aware Refiner (AnoRefiner) that enhances these initial anomaly maps at test time without requiring any labeled samples.

\begin{figure*}
  \centering
  \setlength{\belowcaptionskip}{-7pt}
   \includegraphics[width=1\linewidth]{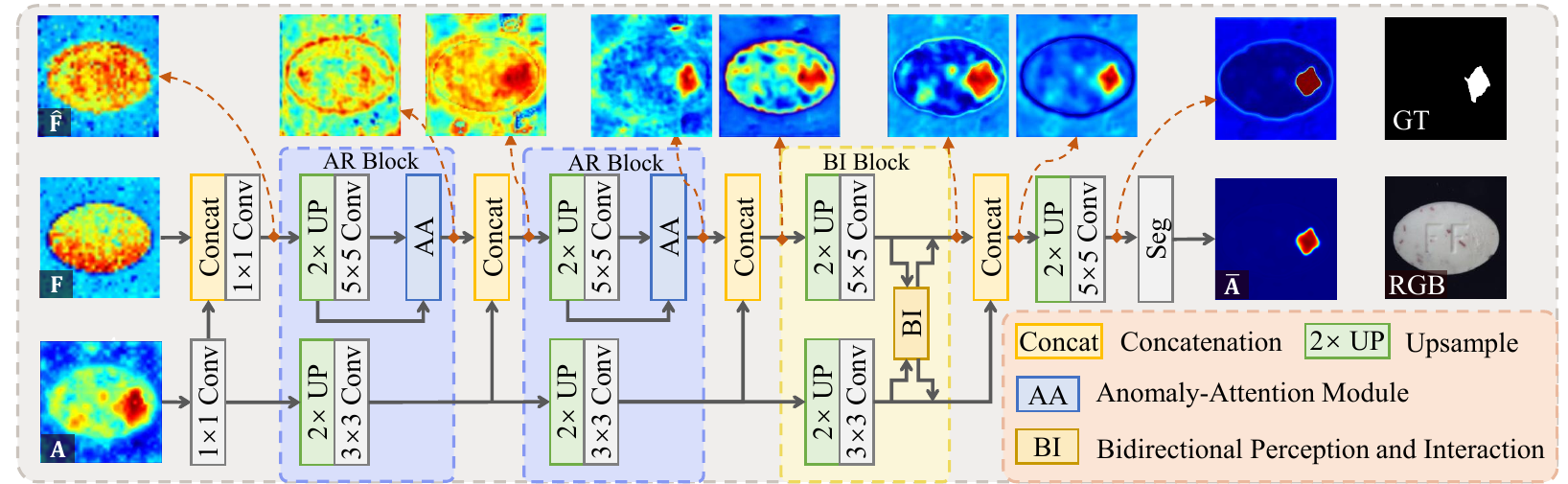}
  \caption{Overview and intermediate results of our Anomaly Refinement Decoder (ARD). The ARD progressively refines image features using anomaly score maps through two Anomaly-aware Refinement (AR) Blocks and one Bidirectional Perception and Interaction (BI) Block. This process enables fine-grained anomaly localization without requiring real anomaly data.}
  \label{fig:pipeline}
\end{figure*}

\subsection{Decoders for Anomaly Refinement}
Some anomaly detection approaches \cite{bae2023pni,yang2023memseg,lu2023hierarchical} often employ decoder-based networks to obtain refined anomaly maps.
For example, RealNet \cite{zhang2024realnet} employs a linear discriminator trained on generated anomaly images to refine features to the pixel level.
DeSTSeg \cite{zhang2023destseg} utilizes a decoder integrated with an Atrous Spatial Pyramid Pooling (ASPP) module \cite{chen2017deeplab}, which is trained on pseudo-anomalies to capture multi-scale contextual information.
DRAEM \cite{zavrtanik2021draem} adopts a U-Net \cite{miccai2015unet} decoder with upsampling for anomaly refinement.
However, these methods depend on pseudo-anomalies that often exhibit a significant gap compared to real anomalies, resulting in inaccurate anomaly localization.
To overcome this limitation, we propose an Anomaly Refinement Decoder (ARD).
Our ARD progressively enhances image features by leveraging the anomaly score map, enabling mutual cross-refinement between both components for precise segmentation.

\subsection{Test Time Training}
Test-time training (TTT) methods \cite{jangtest,liu2021ttt++,sun2020test} fine-tune models during inference and have gained attention, particularly in natural scenes.
While approaches such as TENT \cite{wangtent}, SAR \cite{niutowards}, and MEMO \cite{zhang2022memo} utilize pseudo-labels primarily for entropy minimization or consistency regularization, they focus on classification and offer limited improvements for dense segmentation tasks.
In contrast, we introduce a Progressive Group-wise Test-time Training (PGT) strategy that is specifically designed for anomaly segmentation.
Our PGT trains the decoder incrementally on each group of test images, enabling progressively more accurate anomaly segmentation in subsequent groups.

\section{Method}

In this section, we introduce the Anomaly-aware Refiner (AnoRefiner), it consists of an Anomaly Refinement Decoder (ARD) (see \cref{sec:branch}),  
and a Progressive Group-wise Test Time Training (PGT) strategy (see \cref{Sec:TTT}).

\subsection{Anomaly Refinement Decoder}
\label{sec:branch}

Given the test images $\mathbf{I}=\{I_1,I_2,...,I_N\}$, where $N$ is the image number.
We extract the ViT feature $\mathbf{F} \in \mathbb{R}^{\frac{H}{14} \times \frac{W}{14} \times C}$ of $I_i$, and leverage current ZSAD approaches \cite{li2024musc, zhou2023anomalyclip} to compute the anomaly score of each token of $I_i$, resulting in the anomaly map $\mathbf{A} \in \mathbb{R}^{\frac{H}{14} \times \frac{W}{14} \times 1}$,
which provides a coarse localization of anomalous regions.
To refine $\mathbf{A}$, existing approaches \cite{zavrtanik2021draem} often leverage the synthesis anomalies to train a supervised decoder, while the key issue is that if the feature distribution of the synthesis anomalies is totally different with the real anomalies, the feature may have a lower response in the anomaly region, resulting in the missed detection in the refinement process. 
As shown on the left of \cref{fig:pipeline}, $\textbf{F}$ only exhibits a few responses in the anomaly regions, while $\mathbf{A}$ has a higher anomaly score in the anomaly region.   
Therefore, we propose to effectively enhance the image feature response   
by utilizing the anomaly score map.

Specifically, we first expand $\mathbf{A}$ to 64 channels via a $1\!\times\!1$ convolution, resulting in the initialized auxiliary anomaly score features $\mathbf{\hat{A}}\!\in\!\mathbb{R}^{\frac{H}{14} \times \frac{W}{14} \times 64}$.
Then, we concatenate $\mathbf{F}$ and $\mathbf{\hat{A}}$, 
and fuse them through one $1\!\times\!1$ convolution to obtain the initialized anomaly-aware image feature $\mathbf{\hat{F}}\!\in\!\mathbb{R}^{\frac{H}{14} \times \frac{W}{14} \times 256}$. 

\textbf{Anomaly-aware Refinement Block.}
As shown in Fig. \ref{fig:pipeline}, 
the anomaly-aware refinement (AR) block contains two branches.
Taking $\mathbf{\hat{F}}$ and $\mathbf{\hat{A}}$ as input, we first perform upsampling of them, followed by a $5\times 5$ convolution for the image features to smooth background noise, and a $3\times 3$ convolution for the anomaly score features to suppress interpolation artifacts introduced by the $2\times$ upsampling operation.
Subsequently, the image features are refined by the Anomaly-Attention (AA) module to be more responsive to anomalies.
We perform a two-time stack of the AR Block to progressively enhance the anomaly-aware features, thereby integrating the acquisition of anomaly-aware features with upsampling of the image resolution.
The output of the AA module is concatenated with the anomaly score features.

\begin{figure}
  \centering
  \includegraphics[width=1\linewidth]{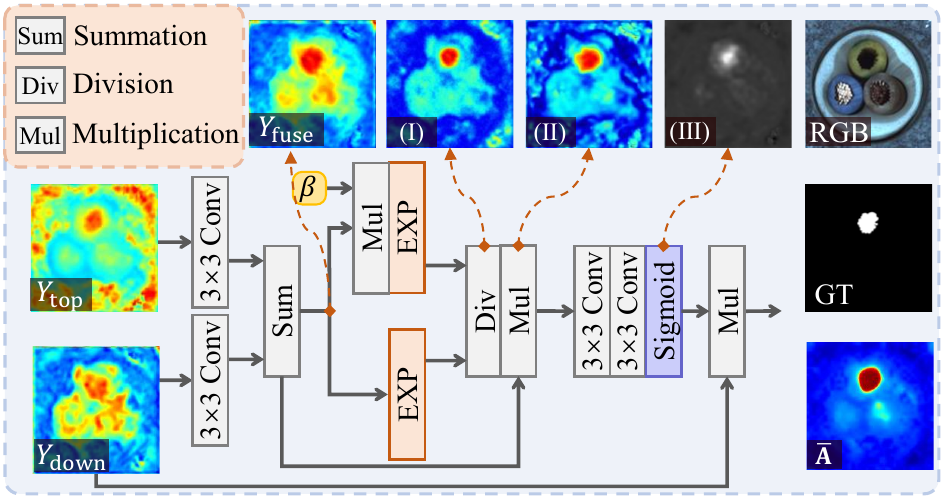}
  \caption{Anomaly-Attention Module architecture and its intermediate feature visualizations.}
  \label{fig:AAM_blocks}
\end{figure}

\textbf{Anomaly-Attention Module.}
Within each AR block, the Anomaly-Attention(AA) module is incorporated, designed specifically to further enhance the anomaly response of the image features.
This module takes the features before ($Y_{\text{top}}$ in \cref{fig:AAM_blocks}) and after ($Y_{\text{down}}$ in \cref{fig:AAM_blocks}) the $5\times 5$ convolution as inputs, and a $3\times 3$ convolution is applied to each of them for dimension adjustment.
The $5\times 5$ convolution reduces background noise in $Y_{\text{top}}$, but may weaken discriminative ability in abnormal regions.
Therefore, we sum $Y_{\text{top}}$ and $Y_{\text{down}}$ to achieve the balanced denoised feature $Y_{\text{fuse}} \in \mathbb{R}^{\frac{H}{7}\times \frac{W}{7}\times 192}$ in \cref{fig:AAM_blocks}.
Subsequently, we calculate the adaptive feature importance weight $\zeta \in \mathbb{R}^{\frac{H}{7}\times \frac{W}{7} \times 192}$  as follows,
\begin{equation}
\begin{gathered}
\zeta =
\frac{\exp({Y_{\text{fuse}}\odot\beta})}
{\exp({Y_{\text{fuse}}})}
= \exp((\beta - \textbf{1})\odot Y_{\text{fuse}})
\end{gathered}
\label{Eq:weight}
\end{equation}
where ${\odot}$ is element-wise multiplication.
$\mathbf{1} \in \mathbb{R}^{\frac{H}{7}\times \frac{W}{7} \times 1}$  is an all-ones matrix, and $\beta \in \mathbb{R}^{\frac{H}{7}\times \frac{W}{7} \times 1}$ is a learnable coefficient matrix, initialized to all ones.
Therefore, the value of $\zeta$ is also all ones initially.
During the learning process, the anomaly tokens in $Y_{\text{fuse}}$ are assigned higher coefficients, and thus their feature values are enlarged through the element-wise multiplication in the numerator of Eq. \eqref{Eq:weight},  while the feature values of the normal tokens are decreased by using lower coefficients.
The $\exp$ operation further strengthens such feature value amplification.
The denominator in Eq. \eqref{Eq:weight} serves as a feature-scale normalizer that preserves the relative magnitude relationships across spatial locations.
We then calculate the weighted features $Y \in \mathbb{R}^{\frac{H}{7}\times \frac{W}{7} \times 192} $ as follows,
\begin{equation}
    Y=Y_{\text{fuse}} \odot \zeta 
\end{equation}
As shown in \cref{fig:AAM_blocks} (II), $Y$ exhibits high responses in the anomaly region and is well separated from the background.
However, our experiments reveal that directly using $Y$ for subsequent anomaly decoding significantly impaired the accuracy, and the reason is that the $\exp$ operation in Eq. \eqref{Eq:weight} changes feature value distribution.
Consequently, instead of directly using $Y$, we convert $Y$ into the anomaly-attention (AA) weight $W \in \mathbb{R}^{\frac{H}{7}\times \frac{W}{7} \times 1} $ using two consecutive $3\times 3$ convolutions followed by a Sigmoid activation.
As shown in \cref{fig:AAM_blocks} (III), AA weight can well attention the anomaly region.
We then perform the element-wise multiplication between $W$ and  $Y_{\text{down}}$ to obtain the output anomaly-attention image feature. 

\begin{figure}
  \centering
  \includegraphics[width=1\linewidth]{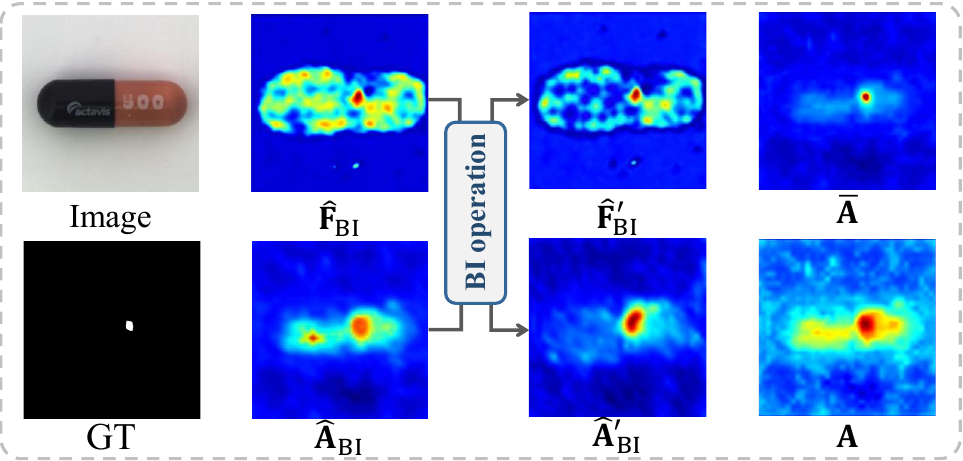}
  \caption{Visualization of feature maps before and after Bidirectional Perception and Interaction operation.}
  \label{fig:BI_blocks}
\end{figure}

\textbf{Bidirectional Perception and Interaction Block.}
As illustrated in \cref{fig:pipeline}, in order to make full use of the anomaly score map $\mathbf{A}$, a series of upsampling operations is applied to $\mathbf{A}$ throughout the refinement stage. 
As we further show in \cref{fig:BI_blocks}, since the anomaly score feature $\mathbf{A}$ is mainly utilized to enhance the image feature, the noise responses in $\mathbf{A}$ cannot be suppressed during the upsampling process. 
In contrast, the image feature $\hat{\mathbf{F}}_{\text{BI}}$ is greatly enhanced through the AA module, and is complementary to the upsampled anomaly score feature $\mathbf{\hat{A}}_{\text{BI}}$.
Therefore, we leverage $\hat{\mathbf{F}}_{\text{BI}}$ and $\mathbf{\hat{A}}_{\text{BI}}$ to cross refine each other, 
and propose the Bidirectional Perception and Interaction (BI) operation as follows:
\begin{equation}
\begin{gathered}
\mathbf{\hat{A}}_{\text{BI}}^{'} = \mathbf{\hat{A}}_{\text{BI}} + \mathrm{Conv}_{3\times3}\!\Bigl(\mathrm{ReLU}\bigl(\mathrm{Conv}_{3\times3}(\mathbf{\hat{F}}_{\text{BI}})\bigr)\Bigr), \\
\mathbf{\hat{F}}_{\text{BI}}^{'} = \mathbf{\hat{F}}_{\text{BI}} + \mathrm{Conv}_{3\times3}\!\Bigl(\mathrm{ReLU}\bigl(\mathrm{Conv}_{3\times3}(\mathbf{\hat{A}}_{\text{BI}})\bigr)\Bigr)
\end{gathered}
\end{equation}
where $\mathbf{\hat{A}}_{\text{BI}}^{'}$ and $\mathbf{\hat{F}}_{\text{BI}}^{'}$ denote $\mathbf{\hat{A}}_{\text{BI}}$ and $\mathbf{\hat{F}}_{\text{BI}}$ after BI operation. $\mathrm{Conv}_{3\times3}$ is a $3\times 3$ convolution, and ReLU is rectified linear unit. We place the BI block at the end of ARD to prevent early fusion from corrupting the $\mathbf{\hat{A}}$ with noisy $\mathbf{\hat{F}}$.

After BI block, the fused feature is passed through the upsample block to reach the original input resolution, and processed by a segmentation head to get the refined anomaly map $\mathbf{\bar{A}} \in \mathbb{R}^{H \times W \times 1}$.

\begin{figure*}
  \centering
  \setlength{\belowcaptionskip}{-7pt}
   \includegraphics[width=1\linewidth]{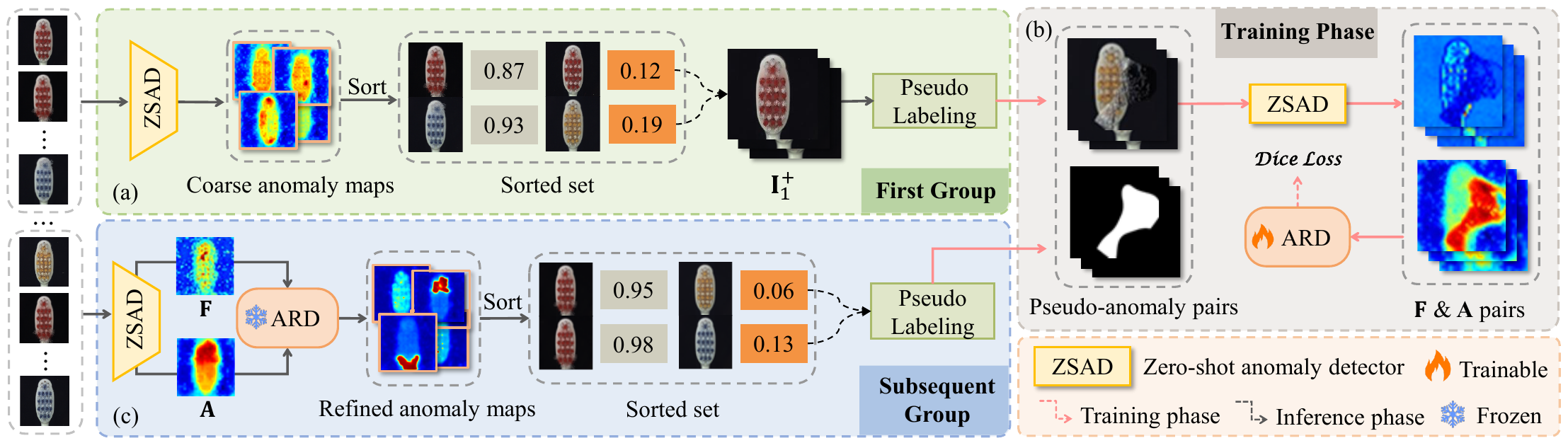}
  \caption{Overview of our PGT. (a) Using ZSAD to produce coarse anomaly map and generate pseudo-anomaly images. (b) Pseudo-anomaly images are used to train the ARD. (c) The trained ARD refines anomaly maps for subsequent groups and generates new pseudo-anomaly images for progressive learning.}
  \label{fig:TTT}
\end{figure*}

\subsection{Progressive Group-wise Test-Time Training}
\label{Sec:TTT}

In this section, we propose a Progressive Group-wise Test-Time Training (PGT) strategy.
As shown in the left of \cref{fig:TTT}, 
the test images is divided into several non-overlapping groups $\{\mathbf{I}_1,\mathbf{I}_2,...,\mathbf{I}_k\}$,
and each group contains $u$ images. 
For each group, PGT performs the following three steps: pseudo-normal samples selection, pseudo-anomaly images synthesis, and testing time training.

\textbf{Pseudo-normal Images Selection.}
As shonw in \cref{fig:TTT} (a),
for each image $I_u \in \mathbf{I_1}$,
we use a ZSAD approach to generate its coarse anomaly map $\mathbf{A}_u$, 
and use the maximum value of $\mathbf{A}_u$ as its image level anomaly score.
According, we select the \textit{top}-$r$ images in $\mathbf{I}_1$ with the lowest anomaly scores as the pseudo-normal samples $\textbf{I}^{+}_1 =\{I^{+}_1,I^{+}_2,...,I^{+}_r\}$.

\textbf{Pseudo-anomaly Images Synthesis.}
Aiming to validate the effectiveness of the proposed anomaly refinement decoder, 
we did not make a special design for the anomaly image synthesis, and just followed existing approaches \cite{zavrtanik2021draem}.
Specifically, we generate a set of random anomalies by leveraging the random masks generated by the Perlin noise, 
and the textures from the DTD dataset ~\cite{cimpoi2014describing}.
These random anomalies are pasted to the pseudo-normal image $\textbf{I}^{+}_{1}$ to synthesize the pseudo-anomaly images $\textbf{I}^P_1 = \{I^p_1, I^p_2, ..., I^p_{r\times l}\}$ with their pixel-level masks $\textbf{M}^P_1 = \{M^p_1, M^p_2, ..., M^p_{r\times l}\}$, where each pseudo-normal image generates $l$ pseudo-anomaly images.

\textbf{Model Training.}
As  shown in \cref{fig:TTT} (b),
we leverage the pseudo-anomaly pairs $(\textbf{I}^P_1,\textbf{M}^P_1)$ to train ARD during test time as follows, 
\begin{align}
\label{eq:Loss2}
\mathcal{L} &= \sum \text{Dice}(\bar{\mathbf{A}}_{i}^P,M^P_{i})
\end{align}
where $\text{Dice}$ is the Dice loss~\cite{milletari2016v}.
As the ARD is not trained before the first group $\mathbf{I}_1$ is coming,  
the coarse anomaly maps in the first group are used directly as the final segmentation result.
In the subsequent group, shown in  \cref{fig:TTT} (c), the ARD trained on preceding groups is leveraged to refine the coarse anomaly map $\mathbf{A}$.
This progressive learning strategy enables the model to steadily improve its representation by incorporating knowledge from previously processed groups, thereby enhancing localization accuracy over time.
Note that, to balance the trade-off between false positives in the coarse anomaly map $\mathbf{A}$ and potential false negatives in the refined anomaly map $\bar{\mathbf{A}}$, we compute their average as the final segmentation result.

\section{Experiments}
\label{sec:experiments}

\begin{table*}[h!]
  \centering
  \setlength{\belowcaptionskip}{-7pt}
  \setlength\tabcolsep{3mm}
    \resizebox{1.0\linewidth}{!}{\begin{tabular}{clccccccc}
    \specialrule{1.5pt}{0pt}{0pt} 
    \textbf{Datasets} & \textbf{Methods} & \textbf{AUROC-cls} & \textbf{F1-max-cls} & \textbf{AP-cls} & \textbf{AUROC-segm} & \textbf{F1-max-segm} & \textbf{AP-segm}\\
    \hline
    \midrule
    \multirow{17}{*}{\textbf{MVTec AD~\cite{bergmann2019mvtec}}} & APRIL-GAN~\cite{chen2023april} & 86.1 & 90.4 & 93.5 & 87.6 & 43.3 & 40.8\\
    ~ & + TIPI~\cite{nguyen2023tipi} & 86.6 & 90.7 & 93.9 & 87.6 & 43.4 & 40.9 \\
    ~ & + SAR~\cite{niutowards} & 86.8 & 90.8 & 94.0 & 87.7 & 43.4 & 40.9 \\
    ~ & + DeYO~\cite{leeentropy} & 87.7 & 90.5 & 94.7 & 87.6 & 43.4 & 40.6 \\
    ~ & + \textbf{Ours} & \textbf{89.2\footnotesize(+3.1)} & \textbf{91.4\footnotesize(+1.0)} & \textbf{95.3\footnotesize(+2.8)} & \textbf{87.8\footnotesize(+0.2)} & \textbf{46.3\footnotesize(+3.0)} & \textbf{44.4\footnotesize(+3.6)} \\
    \cmidrule(lr){2-8}
    ~ & VCP-CLIP~\cite{qu2024vcp} & \textbf{92.1} & 91.7 & 95.5 & 92.0 & 49.4 & 49.1\\
    ~ & + TIPI~\cite{nguyen2023tipi} & 90.0 & 91.8 & 95.6 & 92.1 & 49.3 & 49.1 \\
    ~ & + SAR~\cite{niutowards} & 89.7 & 91.9 & 95.5 & 92.1 & 49.3 & 49.1 \\
    ~ & + DeYO~\cite{leeentropy} & 90.6 & 92.1 & 95.6 & 91.9 & 49.1 & 48.7 \\
    ~ & + \textbf{Ours} & 90.0 & \textbf{92.3\footnotesize(+0.6)} & \textbf{95.7\footnotesize(+0.2)} & \textbf{92.3\footnotesize(+0.3)} & \textbf{52.2\footnotesize(+2.8)} & \textbf{51.6\footnotesize(+2.5)} \\
    \cmidrule(lr){2-8}
    ~ & MuSc~\cite{li2024musc}  & \textbf{97.8} & \textbf{97.5} & \textbf{99.1} & 97.3 & 62.6 & 62.7 \\
    ~ & MuSc*~  & 97.1 & \textbf{97.5} & 98.7 & 97.3 & 61.6 & 61.2 \\
    ~ & + TIPI~\cite{nguyen2023tipi} & 97.6 & 97.2 & 99.0 & 97.1 & 62.2 & 62.4 \\
    ~ & + SAR~\cite{niutowards} & 97.5 & 97.0 & 98.9 & 97.1 & 62.2 & 62.3 \\
    ~ & + DeYO~\cite{leeentropy} & 97.7 & 97.1 & 98.9 & 97.2 & 62.5 & 62.7 \\
    ~ & + \textbf{Ours}  & 97.2 & \textbf{97.5} & 98.7 & \textbf{97.7\footnotesize(+0.4)} & \textbf{64.4\footnotesize(+2.8)} & \textbf{66.3\footnotesize(+5.1)}\\
    \midrule
    \multirow{17}{*}{\textbf{VisA~\cite{zou2022spot}}} & APRIL-GAN~\cite{chen2023april} & 78.0 & 78.7 & 81.4 & 94.2 & 32.3 & 25.7 \\
    ~ & + TIPI~\cite{nguyen2023tipi} & 79.0 & 79.4 & 82.2 & 94.2 & 32.1 & 25.6 \\
    ~ & + SAR~\cite{niutowards} & 79.4 & 79.3 & 82.3 & 94.3 & 32.1 & 25.6 \\
    ~ & + DeYO~\cite{leeentropy} & 79.4 & 79.3 & 82.3 & 94.3 & 32.1 & 25.6 \\
    ~ & + \textbf{Ours} & \textbf{81.4\footnotesize(+3.4)} & \textbf{79.7\footnotesize(+1.0)} & \textbf{86.1\footnotesize(+4.7)} & \textbf{94.8\footnotesize(+0.6)} & \textbf{35.1\footnotesize(+2.7)} & \textbf{30.9\footnotesize(+5.2)} \\
    \cmidrule(lr){2-8}
    ~ & VCP-CLIP~\cite{qu2024vcp} & 83.8 & 81.5 & 87.3 & 95.7 & 34.7 & 31.0 \\
    ~ & + TIPI~\cite{nguyen2023tipi} & 83.6 & 81.5 & 87.4 & 95.8 & 35.2 & 29.7 \\
    ~ & + SAR~\cite{niutowards} & 83.3 & 81.4 & 86.9 & 95.8 & 35.0 & 29.6 \\
    ~ & + DeYO~\cite{leeentropy} & 83.3 & 81.5 & 86.7 & 95.8 & 34.6 & 29.1 \\
    ~ & + \textbf{Ours} & \textbf{84.5\footnotesize(+0.7)} & \textbf{82.0\footnotesize(+0.5)} & \textbf{87.9\footnotesize(+0.6)} & \textbf{96.0\footnotesize(+0.3)} & \textbf{38.1\footnotesize(+3.4)} & \textbf{32.9\footnotesize(+1.9)} \\

    \cmidrule(lr){2-8}
    ~ & MuSc~\cite{li2024musc} & 92.6 & 89.1 & 93.3 & 98.7 & 48.9 & 45.4 \\
    ~ & MuSc*~ & 91.1 & 90.1 & 92.8 & 98.6 & 48.4 & 45.0 \\
    ~ & + TIPI~\cite{nguyen2023tipi} & 92.6 & 89.0 & 93.3 & 98.7 & 48.9 & 45.4 \\
    ~ & + SAR~\cite{niutowards} & \textbf{93.0} & 89.1 & \textbf{93.8} & 98.7 & 49.0 & 45.4 \\
    ~ & + DeYO~\cite{leeentropy} & 92.8 & 89.8 & 93.5 & 98.6 & 48.4 & 44.6 \\
    ~ & + \textbf{Ours} & 91.9 & \textbf{90.3\footnotesize(+0.2)} & 92.6 & \textbf{98.9\footnotesize(+0.3)} & \textbf{51.3\footnotesize(+2.9)} & \textbf{48.0\footnotesize(+3.0)} \\
    \specialrule{1.5pt}{0pt}{0pt}
  \end{tabular}}
  \caption{Quantitative comparisons on the MVTec AD and VisA dataset. We report the average performance over 5 random seeds. Bold indicates the best performance. * indicates that the identical grouping method from our default settings is used for grouping processing.}
  \label{tab:zeroshotplus}
\end{table*}

\textbf{Implementation Details.}
In this paper, we integrate our method with existing zero-shot anomaly detection approaches \cite{chen2023april, qu2024vcp, li2024musc} to refine their generated anomaly maps.
The feature extractor and input resolution strictly follow the settings of the respective zero-shot method.
In our implementation, all images from a category are uniformly partitioned into a set of groups, each containing $u=30$ images.
Every group is trained for 20 epochs using SGD \cite{robbins1951sgd} with a learning rate of 0.001.
We set the number of pseudo-normal images $r$ to 5, and generate $l\!=\!4$ pseudo-anomaly images for each pseudo-normal image.

\noindent\textbf{Datasets.}
We conduct comprehensive experiments to evaluate our AnoRefiner's performance in industrial anomaly classification and segmentation, we use two popular industrial image datasets: MVTec AD~\cite{bergmann2019mvtec} and VisA~\cite{zou2022spot}.
The MVTec AD dataset contains high-resolution RGB images ranging from $700^2$ to $1024^2$ pixels, encompassing 10 object categories and 5 texture categories.
The VisA dataset offers complementary challenges with its 1000×1500 pixel RGB images spanning 12 objects across 3 distinct domains.
Both datasets contain a mix of normal and anomalous images.

\noindent\textbf{Evaluation Metrics.}
For image-level anomaly classification, we use the Area Under the Receiver Operating Characteristic curve (AUROC), Average Precision (AP) and the maximum F1-score (F1-max).
For pixel-level anomaly segmentation, we adopt the corresponding pixel-wise AUROC, AP, and F1-max.
These metrics collectively evaluate the detection and localization performance.

\noindent\textbf{Baselines.}
We integrate our approach with some zero-shot anomaly detection methods, including APRIL-GAN~\cite{chen2023april}, VCP-CLIP~\cite{qu2024vcp}, and MuSc~\cite{li2024musc}, and compare against test-time training methods such as TIPI~\cite{nguyen2023tipi}, SAR~\cite{niutowards}, and DeYO~\cite{leeentropy}.
Notably, the unmeasured metrics in these papers are reproduced by using the official implementation code. All experiments follow the same implementation details to ensure a fair comparison.
Specifically, we keep the same grouping strategy for processing each product’s data in groups, collecting predictions and features, computing model uncertainty (e.g., entropy and PLPD from DeYO~\cite{leeentropy}), and using these scores as the loss to update the BatchNorm/LayerNorm parameters of corresponding backbone blocks during test time.

\subsection{Quantitative and Qualitative Results}
\noindent\textbf{Quantitative Comparison.}
We evaluate our AnoRefiner for zero-shot anomaly segmentation on MVTec AD and VisA in Tab. \ref{tab:zeroshotplus}.
Our method brings consistent improvements across different frameworks, particularly in segmentation metrics.
While anomaly classification gains are modest, this is consistent with AnoRefiner's focus on refining anomaly maps at the patch level rather than extracting image-level features.
On MVTec AD, AnoRefiner raises F1-max of VCP-CLIP by 2.8\%.
When integrated with MuSc, segmentation AUROC improves by 0.4\% and AP by 5.1\%.
Similarly, on VisA, VCP-CLIP with AnoRefiner achieves a 3.4\% gain in F1-max, while MuSc sees a 3.0\% rise in AP.
These results confirm that AnoRefiner effectively mitigates the imprecision of patch-level features in zero-shot anomaly detection methods, delivering refined, pixel-aware anomaly maps across diverse backends.
Furthermore, comparisons with three natural-scene test-time training methods confirm that our AnoRefiner achieves superior segmentation performance,
demonstrating its effectiveness across diverse industrial zero-shot anomaly detection frameworks.

\begin{figure*}[t]
  \centering
  \setlength{\belowcaptionskip}{-7pt}
  \includegraphics[width=1\linewidth]{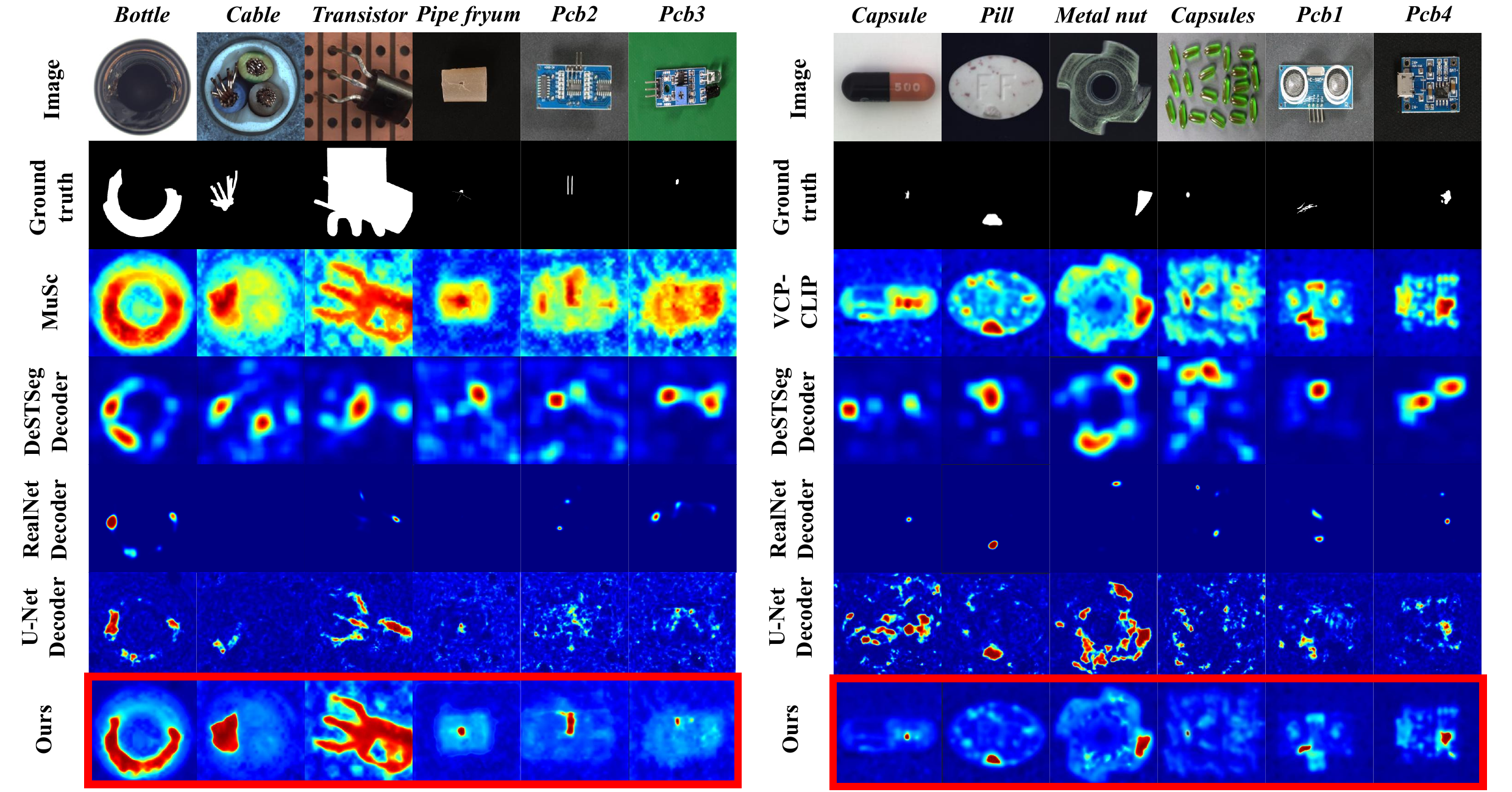}
   \caption{Visualization of anomaly segmentation results on MVTec AD and VisA benchmarks. Row 4-6 present results after replacing our Anomaly Refinement Decoder with the decoder architectures from U‑Net, DeSTSeg, and RealNet, respectively.}
   \label{fig:experiments}
\end{figure*}

\noindent\textbf{Qualitative comparison.}
In \cref{fig:experiments}, we visualize the anomaly segmentation results of MuSc \cite{li2024musc} and VCP-CLIP \cite{qu2024vcp} integrated with our AnoRefiner framework.
Compared to the original anomaly maps from zero-shot methods, our refined maps (red boxes) exhibit significantly reduced background noise.
For structurally complex objects such as the PCB, baseline methods tend to produce false positives in normal regions.
After refinement, these false positives are substantially suppressed, leading to more precise segmentation boundaries.
To further validate the effectiveness of our Anomaly Refinement Decoder, we replace it with decoders from U-Net \cite{miccai2015unet}, DeSTSeg \cite{zhang2023destseg}, and RealNet \cite{zhang2024realnet}.
These decoders lack explicit guidance from the anomaly map, resulting in inferior localization performance, particularly in zero-shot anomaly detection methods where no real anomaly images are available for training.

\begin{table}[t]
\resizebox{\linewidth}{!}{
  \centering
  \setlength{\belowcaptionskip}{-10pt}
  \setlength{\tabcolsep}{2mm}{
  \begin{tabular}{lccccccc}
    \toprule
       \multirow{2}{*}{\textbf{Baselines}} &
        \multicolumn{3}{c}{\textbf{Modules}}&
        \multicolumn{2}{c}{\textbf{MVTec AD}} &
        \multicolumn{2}{c}{\textbf{VisA}} \\
    \cmidrule(r){2-4} \cmidrule(l){5-8}
      & \textbf{AnoB} & \textbf{AA}&\textbf{BI}& \textbf{Image-level} & \textbf{Pixel-level} 
& \textbf{Image-level} & \textbf{Pixel-level}
\\
    \midrule
    \multirow{4}{*}{VCP-CLIP~}& &  && (91.5,95.3) & (50.2,49.6) 
& (81.6,86.9)& (35.2,31.5)\\
 & \checkmark& & & (91.5,95.4)& (51.3,50.8)& (81.6,87.1)&(36.6,31.8)\\
    & \checkmark &  \checkmark&
& (91.8,95.6)& (51.8,51.0)& (81.8,87.7)& (37.4,32.1)\\ 
    & \checkmark  & \checkmark  &\checkmark&(\textbf{92.3},\textbf{95.7})& (\textbf{52.2},\textbf{51.6})& (\textbf{82.0},\textbf{87.9)}& (\textbf{38.1},\textbf{32.9)}\\
    \midrule
    \multirow{4}{*}{MuSc~}& &  && (97.2,98.3)& (62.1,62.8)& (90.1,92.3)& (49.1,46.0)\\
 & \checkmark& & & (97.2,98.6)& (63.6,64.7)& (90.1,92.4)&(50.2,46.9)\\
    & \checkmark &  \checkmark&
& (97.3,98.6) & (63.8,65.5)& (90.0,92.4)& (50.6,47.5)\\ 
    & \checkmark  & \checkmark  &\checkmark& (\textbf{97.5},\textbf{98.7})& (\textbf{64.4},\textbf{66.3})& (\textbf{90.3},\textbf{92.6})& (\textbf{51.3},\textbf{48.0})\\
    \bottomrule
  \end{tabular}
  }
  }
    \caption{Ablation study of main components of Anomaly Refinement Decoder in image-level and pixel-level (F1-max, AP) on MVTec AD and VisA datasets.}
    \label{tab:BSR}
\end{table}

\begin{table}
\resizebox{\linewidth}{!}{
\centering
\setlength{\belowcaptionskip}{-8pt}
\begin{tabular}{llcccc}
\toprule
            \multirow{2}{*}{\textbf{Baselines}} &
            \multirow{2}{*}{\textbf{Module}} &
            \multicolumn{2}{c}{\textbf{MVTec AD}} &
            \multicolumn{2}{c}{\textbf{VisA}} \\
            \cmidrule(r){3-4} \cmidrule(l){5-6}
            & & \textbf{Image-level}& \textbf{Pixel-level} & \textbf{Image-level}& \textbf{Pixel-level} \\
            \midrule
            \multirow{2}{*}{APRIL-GAN} & SA & (91.3,95.1)& (44.9,43.0)& (79.3,85.2)& (34.4,30.3)\\ 
            & AA (Ours) & (\textbf{91.4},\textbf{95.3})& (\textbf{46.3},\textbf{44.4})& (\textbf{79.7},\textbf{86.1})& (\textbf{35.1},\textbf{30.9})\\
            \midrule
            \multirow{2}{*}{VCP-CLIP} & SA & (91.7,95.2) & (51.2,49.8) & (81.6,87.6) & (37.2,32.3)\\ 
            & AA (Ours) & (\textbf{92.3},\textbf{95.7})& (\textbf{52.2},\textbf{51.6})& (\textbf{82.0},\textbf{87.9})& (\textbf{38.1},\textbf{32.9})\\
            \midrule
            \multirow{2}{*}{MuSc}& SA & (97.3,98.6) & (62.9,63.8) & (90.0,92.2) & (49.3,46.5) \\ 
            & AA (Ours) & (\textbf{97.5},\textbf{98.7})& (\textbf{64.4},\textbf{66.3})& (\textbf{90.3},\textbf{92.6})& (\textbf{51.3},\textbf{48.0})\\
            
\bottomrule
        \end{tabular}
}
\caption{Ablation study of our Anomaly-Attention (AA) in image-level and pixel-level (F1-max, AP) on MVTec AD and VisA.}
\label{tab:SE}
\end{table}

\subsection{Ablation study}
\label{ablation_study}

\textbf{Ablation study of Anomaly Refinement Decoder.}
\cref{tab:BSR} presents the ablation study of the main components in our Anomaly Refinement Decoder: the auxiliary anomaly score map branch (AnoB), the Anomaly-Attention (AA) module, and the Bidirectional Perception and Interaction (BI) block.
The AnoB branch uses the auxiliary anomaly score map as guidance to reduce missed detections, which improves average pixel-level AP by 1.3\% on MVTec AD.
The AA and BI modules further enhance the discriminability between normal and abnormal regions, contributing additional gains of 1.2\% on average in pixel-level AP.
As the decoder is designed primarily to refine pixel-wise segmentation, its impact on image-level performance is small.

\noindent\textbf{Ablation study of Anomaly-Attention.}
We conduct experiments to replace the proposed Anomaly-Attention (AA) with vanilla spatial attention (SA) in \cref{tab:SE}.
Our module first separates anomaly regions from the background, which facilitates the subsequent learning of attention weights.
This design yields a gain of 0.6\% in image-level F1-max and 2.5\% in pixel-level AP on the MVTec AD dataset.

\begin{table}[t]
\resizebox{\linewidth}{!}{
\scriptsize
  \centering
  \setlength{\tabcolsep}{3mm}{
  \begin{tabular}{lcccccc}
    \toprule
       \multirow{3}{*}{\textbf{Baselines}} &
         \multirow{2}{*}{\textbf{Contamination}} &
        \multicolumn{2}{c}{\textbf{MVTec AD}} &
        \multicolumn{2}{c}{\textbf{VisA}} \\
    \cmidrule(r){3-4} \cmidrule(l){5-6}
      & \textbf{Ratio} & \textbf{Image-level}& \textbf{Pixel-level} & \textbf{Image-level}& \textbf{Pixel-level} \\
    \midrule
    \multirow{3}{*}{VCP-CLIP~}& Ours &(\textbf{92.3},\textbf{95.7})& (\textbf{52.2},\textbf{51.6})& (\textbf{82.0},87.9)& (\textbf{38.1},\textbf{32.9})\\
    & 20\% & (92.2,95.5) & (51.8,51.0) & (\textbf{82.0},87.8) & (37.7,32.6) \\
    & 40\% & (92.0,95.3) & (51.5,50.8) & (81.9,87.8) & (37.6,32.7) \\
    \midrule
    \multirow{3}{*}{MuSc~}& Ours & (\textbf{97.5},\textbf{98.7}) & (\textbf{64.4},\textbf{66.3}) & (\textbf{90.3},\textbf{92.6}) & (\textbf{51.3},\textbf{48.0})\\
    & 20\% & (97.3,98.6) & (64.0,66.1) & (90.0,92.5) & (51.1,47.7) \\
    & 40\% & (97.3,98.5) & (63.7,65.7) & (90.0,92.5) & (50.8,47.3) \\
    \bottomrule
  \end{tabular}
  }}
    \caption{Ablation of different contamination ratios of pseudo-normal images in image-level and pixel-level (F1-max, AP).}
    \label{tab:contami_normal}
\end{table}

\noindent\textbf{Discussion of Pseudo-Normal Images Selection.} 
\cref{tab:combined} presents the ablation study on the number of pseudo-normal images $r$.
A small $r$ yields fewer training images, resulting in insufficient training and reduced diversity in the normal pattern representation, which in turn degrades performance.
In contrast, a large $r$ may increase the risk of including anomaly images in the pseudo-normal images.
However, as our decoder uses the anomaly map as a refinement guide, increasing $r$ has a small impact on metrics (less than 1\%).
We further simulate contamination by deliberately introducing real anomaly images into the pseudo-normal images in \cref{tab:contami_normal}.
Under the setting of a 40\% contamination ratio, pixel-level anomaly localization performance drops by only 0.8\% AP on MVTec AD and 0.7\% on VisA, while image-level detection metrics decrease by at most 0.4\%.
These results demonstrate the robustness of our approach to potential anomalies in the pseudo-normal images.

\begin{table}[t]
\resizebox{\linewidth}{!}{
\scriptsize
  \centering
  \setlength{\tabcolsep}{3mm}{
  \begin{tabular}{llccccc}
    \toprule
       \multirow{2}{*}{\textbf{Baselines}} &
         \multirow{2}{*}{\textbf{Strategy}} &
        \multicolumn{2}{c}{\textbf{MVTec AD}} &
        \multicolumn{2}{c}{\textbf{VisA}} \\
    \cmidrule(r){3-4} \cmidrule(l){5-6}
      &  & \textbf{Image-level}& \textbf{Pixel-level} & \textbf{Image-level}& \textbf{Pixel-level} \\
    \midrule
    \multirow{3}{*}{VCP-CLIP~}
    & (a) Perlin\&DTD &(92.3,95.7)& (52.2,51.6)& (82.0,87.9)& (38.1,32.9)\\
    & (b) CutPaste \cite{cvpr2021cutpaste} & (92.1,95.4) & (51.9,51.3) & (81.9,87.9) & (37.8,32.7) \\
    & (c) SeaS \cite{dai2025seas} & (92.4,95.7) & (53.5,52.7) & (82.2,88.0) & (39.0,34.2) \\
    \midrule
    \multirow{3}{*}{MuSc~}
    & (a) Perlin\&DTD & (97.5,98.7) & (64.4,66.3) & (90.3,92.6) & (51.3,48.0)\\
    & (b) CutPaste \cite{cvpr2021cutpaste} & (97.4,98.5) & (64.0,66.1) & (90.3,92.7) & (51.0,47.5) \\
    & (c) SeaS \cite{dai2025seas} & (97.6,98.8) & (65.9,67.6) & (90.5,92.7) & (51.8,48.7)\\
    \bottomrule
  \end{tabular}
  }}
    \caption{Ablation study of different anomaly synthesis strategies in image-level and pixel-level (F1-max, AP).}
    \label{tab:syn_strategy}
\end{table}

\noindent\textbf{Discussion of Pseudo-Anomaly Images Synthesis.}
For anomaly synthesis, we primarily adopt the DTD dataset to obtain abnormal textures.
In \cref{tab:syn_strategy}, we also evaluate two alternative strategies: (b) CutPaste \cite{cvpr2021cutpaste}, which extracts abnormal textures from other pseudo-normal images, and (c) SeaS \cite{dai2025seas}, which learns to simulate anomalies from real anomaly images.
As SeaS utilizes real anomaly data, it achieves better performance.
The other two strategies (a) and (b) have a marginal impact on results (within 0.5\%).
In \cref{tab:combined}, we ablate the number of pseudo-anomalies per normal image $l$.
Similar to the parameter $r$, a small $l$ leads to insufficient training due to limited anomaly images, while a large $l$ costs a longer training time.
We choose $l=4$ as a balanced trade-off.

\begin{table}
\resizebox{\linewidth}{!}{
  \centering
  \setlength{\tabcolsep}{2mm}{
  \begin{tabular}{cccccccc}
    \toprule
    \multirow{2}{*}{\textbf{Methods}} & \multirow{2}{*}{$l/r$} & \multicolumn{1}{c}{\multirow{2}{*}{\textbf{Num}}} & \multicolumn{2}{c}{\textbf{MVTec AD}} & \multicolumn{2}{c}{\textbf{VisA}} \\
    \cmidrule(lr){4-5} \cmidrule(lr){6-7}
    & & & \textbf{Image-level} & \textbf{Pixel-level} & \textbf{Image-level} & \textbf{Pixel-level} \\
    \midrule
    \multirow{7}{*}{VCP-CLIP~}& \multirow{4}{*}{$r$}& 3 & (91.9,95.2)&(50.8,49.9)&(81.4,87.3)&(36.8,30.9)\\
    & & 4 & (92.2,95.5)&(52.0,51.2)&(81.8,87.7)&(37.5,32.6) \\
    & & 5 & (\textbf{92.3},95.7)& (52.2,51.6)& (\textbf{82.0},\textbf{87.9})& (38.1,32.9) \\
    & & 6 & (\textbf{92.3},\textbf{95.8})&(\textbf{52.6},\textbf{52.2})&(\textbf{82.0},87.8)&(\textbf{38.4},\textbf{33.1}) \\
    \cmidrule(lr){2-7}
     &\multirow{3}{*}{$l$} & 2 & (92.0,95.5)&(50.6,49.8)&(81.7,87.3)&(37.0,31.7) \\
    & & 4 & (\textbf{92.3},\textbf{95.7})& (52.2,51.6)& (\textbf{82.0},\textbf{87.9})& (\textbf{38.1},\textbf{32.9}) \\
    & & 6 & (\textbf{92.3},95.6)&(\textbf{52.5},\textbf{52.2})&(81.9,87.7)&(\textbf{38.7},\textbf{33.3}) \\
    \midrule
    \multirow{7}{*}{MuSc~} & \multirow{4}{*}{$r$}& 3 & (97.3,98.5)& (63.8,65.5) & (89.8,92.1) & (50.5,47.1) \\
    & & 4 & (\textbf{97.5},\textbf{98.7})& (64.2,66.0) & (90.1,92.4) & (50.9,47.8) \\
    & & 5 & (\textbf{97.5},\textbf{98.7})&(64.4,66.3)&(\textbf{90.3},\textbf{92.6})&(51.3,48.0) \\
    & & 6 & (97.4,98.6)& (\textbf{64.6},\textbf{66.9}) & (90.2,92.4) & (\textbf{51.9},\textbf{48.6})\\
    \cmidrule(lr){2-7}
     &\multirow{3}{*}{$l$} & 2 & (97.3,98.5)&(63.4,65.2)&(90.0,92.3)&(50.1,46.7) \\
    & & 4 & (\textbf{97.5},\textbf{98.7})&(64.4,66.3)&(\textbf{90.3},\textbf{92.6})&(51.3,48.0) \\
    & & 6 &  (97.2,98.4)&(\textbf{64.7},\textbf{66.8})&(90.1,92.5)&(\textbf{51.9},\textbf{48.7})  \\
    \bottomrule
  \end{tabular}
  }}
  \caption{Sensitivity of pseudo-normal image number $r$ and pseudo-anomalies number $l$ in image-level and pixel-level (F1-max, AP) on MVTec AD and VisA datasets.}
  \label{tab:combined}
\end{table}

\begin{table}[t]
\resizebox{\linewidth}{!}{
\scriptsize
  \centering
  \setlength{\tabcolsep}{3mm}{
  \begin{tabular}{llccccc}
    \toprule
       \multirow{2}{*}{\textbf{Baselines}} &
         \multirow{2}{*}{\textbf{Setting}} &
        \multicolumn{2}{c}{\textbf{MVTec AD}} &
        \multicolumn{2}{c}{\textbf{VisA}} \\
    \cmidrule(r){3-4} \cmidrule(l){5-6}
      &  & \textbf{Image-level}& \textbf{Pixel-level} & \textbf{Image-level}& \textbf{Pixel-level} \\
    \midrule
    \multirow{4}{*}{VCP-CLIP~}
    & (a) one group & (92.1,95.3) & (51.8,51.0) & (81.5,87.6) & (37.6,32.2) \\
    & (b) same anomalies & (92.2,95.3) & (52.0,51.0) & (81.7,87.9) & (37.1,31.9) \\
    & (c) same normalities & (91.9,95.1) & (51.6,50.6) & (81.3,87.5) & (37.2,31.9) \\
    & (d) Ours & (\textbf{92.3},\textbf{95.7})& (\textbf{52.2},\textbf{51.6})& (\textbf{82.0},\textbf{87.9})& (\textbf{38.1},\textbf{32.9}) \\
    \midrule
    \multirow{4}{*}{MuSc~}
    & (a) one group &  (97.3,98.5) & (63.6,65.5) & (90.1,92.2) & (50.0,46.6) \\
    & (b) same anomalies &  (97.3,98.5) & (64.0,65.5) & (\textbf{90.3},92.5) & (50.9,47.5) \\
    & (c) same normalities & (97.1,98.3) & (63.4,65.1) & (89.9,92.1) & (49.6,46.3)\\
    & (d) Ours & (\textbf{97.5},\textbf{98.7}) & (\textbf{64.4},\textbf{66.3}) & (\textbf{90.3},\textbf{92.6}) & (\textbf{51.3},\textbf{48.0}) \\
    \bottomrule
  \end{tabular}
  }}
    \caption{Ablation study of the progressive training process in image-level and pixel-level (F1-max, AP).}
    \label{tab:continue}
\end{table}

\noindent\textbf{Discussion of the Progressive Training.}
\cref{tab:continue} presents an ablation study on our progressive training process using four configurations:
(a) training ARD only on the first group,
(b) using identical pseudo-anomalies across all groups,
(c) sharing pseudo-normal images from the first group but generating different pseudo-anomalies per group,
and (d) our default setup where each group independently selects pseudo-normal images and synthesizes unique pseudo-anomalies.
The improvement over (a) validates the necessity of training across all groups.
Our method (d) consistently outperforms (c) across metrics, indicating that diverse pseudo-normal images from different groups help learn a more comprehensive normal representation.
It also surpasses (b) by 0.7\% in pixel-level AP, underscoring the benefit of pseudo-anomaly diversity.

\section{Conclusion}
This paper presents AnoRefiner, a framework that refines the coarse anomaly maps produced by zero-shot anomaly detection (ZSAD) methods into precise pixel-level segmentations.
AnoRefiner integrates an Anomaly Refinement Decoder (ARD) and a Progressive Group-wise Test-time Training (PGT) strategy, operating without relying on the quality of labeled anomaly data.
The ARD utilizes an Anomaly-Attention module and a Bidirectional Interaction operation to suppress background noise and enhance anomaly cues.
The PGT strategy adapts the model to groups of test images, simulating the continuous flow of industrial production for progressive improvement.

\section{Acknowledgments}
\noindent This work was supported by the National Natural Science Foundation of China under Grant No.62176098. The computation is completed in the HPC Platform of Huazhong University of Science
 and Technology.

{
    \small
    \bibliographystyle{ieeenat_fullname}
    \bibliography{main}

\begin{thebibliography}{15}
\providecommand{\natexlab}[1]{#1}
\providecommand{\url}[1]{\texttt{#1}}
\expandafter\ifx\csname urlstyle\endcsname\relax
  \providecommand{\doi}[1]{doi: #1}\else
  \providecommand{\doi}{doi: \begingroup \urlstyle{rm}\Url}\fi

\bibitem[Bergmann et~al.(2019)Bergmann, Fauser, Sattlegger, and Steger]{appbergmann2019mvtec}
Paul Bergmann, Michael Fauser, David Sattlegger, and Carsten Steger.
\newblock Mvtec ad--a comprehensive real-world dataset for unsupervised anomaly detection.
\newblock In \emph{Proceedings of the IEEE/CVF Conference on Computer Vision and Pattern Recognition}, pages 9592--9600, 2019.

\bibitem[Chen et~al.(2023)Chen, Han, and Zhang]{appchen2023april}
Xuhai Chen, Yue Han, and Jiangning Zhang.
\newblock April-gan: A zero-/few-shot anomaly classification and segmentation method for cvpr 2023 vand workshop challenge tracks 1\&2: 1st place on zero-shot ad and 4th place on few-shot ad.
\newblock \emph{arXiv preprint arXiv:2305.17382}, 2023.

\bibitem[Deng et~al.(2009)Deng, Dong, Socher, Li, Li, and Fei-Fei]{appdeng2009imagenet}
Jia Deng, Wei Dong, Richard Socher, Li-Jia Li, Kai Li, and Li Fei-Fei.
\newblock Imagenet: A large-scale hierarchical image database.
\newblock In \emph{2009 IEEE conference on Computer Vision and Pattern Recognition}, pages 248--255. Ieee, 2009.

\bibitem[Dosovitskiy(2020)]{appdosovitskiy2020image}
Alexey Dosovitskiy.
\newblock An image is worth 16x16 words: Transformers for image recognition at scale.
\newblock In \emph{The Ninth International Conference on Learning Representations}, 2020.

\bibitem[He et~al.(2025)He, Cao, Peng, and Xie]{apphe2025rareclip}
Jianfang He, Min Cao, Silong Peng, and Qiong Xie.
\newblock Rareclip: Rarity-aware online zero-shot industrial anomaly detection.
\newblock In \emph{Proceedings of the IEEE/CVF International Conference on Computer Vision}, pages 24478--24487, 2025.

\bibitem[He et~al.(2016)He, Zhang, Ren, and Sun]{appcvpr2016resnet}
Kaiming He, Xiangyu Zhang, Shaoqing Ren, and Jian Sun.
\newblock Deep residual learning for image recognition.
\newblock In \emph{Proceedings of the IEEE conference on Computer Vision and Pattern Recognition}, pages 770--778, 2016.

\bibitem[Li et~al.(2024)Li, Huang, Xue, and Zhou]{appli2024musc}
Xurui Li, Ziming Huang, Feng Xue, and Yu Zhou.
\newblock Musc: Zero-shot industrial anomaly classification and segmentation with mutual scoring of the unlabeled images.
\newblock In \emph{The Twelfth International Conference on Learning Representations}, 2024.

\bibitem[Ma et~al.(2025)Ma, Zhang, Yao, Tang, Wu, Li, Yan, Jiang, and Zhou]{appma2025aa}
Wenxin Ma, Xu Zhang, Qingsong Yao, Fenghe Tang, Chenxu Wu, Yingtai Li, Rui Yan, Zihang Jiang, and S~Kevin Zhou.
\newblock Aa-clip: Enhancing zero-shot anomaly detection via anomaly-aware clip.
\newblock In \emph{Proceedings of the Computer Vision and Pattern Recognition Conference}, pages 4744--4754, 2025.

\bibitem[Qu et~al.(2024)Qu, Tao, Prasad, Shen, Zhang, Gong, and Ding]{appqu2024vcp}
Zhen Qu, Xian Tao, Mukesh Prasad, Fei Shen, Zhengtao Zhang, Xinyi Gong, and Guiguang Ding.
\newblock Vcp-clip: A visual context prompting model for zero-shot anomaly segmentation.
\newblock In \emph{European Conference on Computer Vision}, pages 301--317, 2024.

\bibitem[Qu et~al.(2025)Qu, Tao, Gong, Qu, Chen, Zhang, Wang, and Ding]{appqu2025bayesian}
Zhen Qu, Xian Tao, Xinyi Gong, Shichen Qu, Qiyu Chen, Zhengtao Zhang, Xingang Wang, and Guiguang Ding.
\newblock Bayesian prompt flow learning for zero-shot anomaly detection.
\newblock In \emph{Proceedings of the Computer Vision and Pattern Recognition Conference}, pages 30398--30408, 2025.

\bibitem[Robbins and Monro(1951)]{approbbins1951sgd}
Herbert Robbins and Sutton Monro.
\newblock A stochastic approximation method.
\newblock \emph{The Annals of Mathematical Statistics}, pages 400--407, 1951.

\bibitem[Ronneberger et~al.(2015)Ronneberger, Fischer, and Brox]{appmiccai2015unet}
Olaf Ronneberger, Philipp Fischer, and Thomas Brox.
\newblock U-net: Convolutional networks for biomedical image segmentation.
\newblock In \emph{International Conference on Medical Image Computing and Computer-assisted Intervention}, pages 234--241. Springer, 2015.

\bibitem[Zhang et~al.(2023)Zhang, Li, Li, Huang, Shan, and Chen]{appzhang2023destseg}
Xuan Zhang, Shiyu Li, Xi Li, Ping Huang, Jiulong Shan, and Ting Chen.
\newblock Destseg: Segmentation guided denoising student-teacher for anomaly detection.
\newblock In \emph{Proceedings of the IEEE/CVF Conference on Computer Vision and Pattern Recognition}, pages 3914--3923, 2023.

\bibitem[Zhang et~al.(2024)Zhang, Xu, and Zhou]{appzhang2024realnet}
Ximiao Zhang, Min Xu, and Xiuzhuang Zhou.
\newblock Realnet: A feature selection network with realistic synthetic anomaly for anomaly detection.
\newblock In \emph{Proceedings of the IEEE/CVF Conference on Computer Vision and Pattern Recognition}, pages 16699--16708, 2024.

\bibitem[Zou et~al.(2022)Zou, Jeong, Pemula, Zhang, and Dabeer]{appzou2022spot}
Yang Zou, Jongheon Jeong, Latha Pemula, Dongqing Zhang, and Onkar Dabeer.
\newblock Spot-the-difference self-supervised pre-training for anomaly detection and segmentation.
\newblock In \emph{European Conference on Computer Vision}, pages 392--408. Springer, 2022.

\end{thebibliography}
}

\appendix

\maketitlesupplementary

\section*{Overview}
In this appendix, we provide additional descriptions of the following contents:
\begin{itemize}
    
    \item The implementation details about our experiment settings (\cref{app:implementation_details}).
    
    \item Discussion of the general compatibility with CNN (\cref{app:CNN}).

    \item Discussion of the summation of results from the ZSAD methods and AnoRefiner(\cref{app:sum}).

    \item Discussion on the different group size in PGT (\cref{app:group}).

    \item Detail ablation study of Anomaly-Attention module (\cref{app:AAM}).

    \item Detail analysis of using different decoder (\cref{app:decoder}).

    \item More comparison of different ZSAD methods (\cref{app:ZSAD}).

    \item More detailed qualitative results
    (\cref{app:qualitative}).
    
    \item More detailed quantitative results (\cref{app:quantitative}).

\end{itemize}

\section{Additional implementation details}
\label{app:implementation_details}
In this paper, we integrate our method with existing zero-shot anomaly detection approaches \citeapp{appchen2023april, appqu2024vcp, appli2024musc} to refine their generated anomaly maps.
The feature extractor and input resolution strictly follow the settings of the respective zero-shot method.
In our implementation, all images from a category are uniformly partitioned into a set of groups, each containing $u=30$ images.
Every group is trained for 20 epochs using SGD \citeapp{approbbins1951sgd} with a learning rate of 0.001.
We set the number of pseudo-normal images $r$ to 5, and generate $l\!=\!4$ pseudo-anomaly images for each pseudo-normal image.
Current zero-shot anomaly detection methods often use the 6, 12, 18, and 24 layers of ViT-L/14-336 \citeapp{appdosovitskiy2020image} to detect anomalies, therefore we average these features as the input of our ARD.
Since the MVTec AD dataset contains only 42 images of the toothbrush product, we split this class into two groups, each consisting of 21 images.

\section{Discussion of the general compatibility with CNN.}
\label{app:CNN}
To evaluate compatibility with CNN-based architectures, we integrate our decoder into MuSc~\citeapp{appli2024musc}, which uses a ResNet-50 backbone~\citeapp{appcvpr2016resnet} pre-trained on ImageNet~\citeapp{appdeng2009imagenet}.
All the input images are resized to a $512\times512$ resolution.
Since the RsCIN module in MuSc is specific to the class token of Vision Transformers (ViTs), we remove it when using CNN features.
Features from the second and third ResNet-50 layers are aggregated as input to MuSc and our AnoRefiner.

The results are shown in Table \ref{tab:cnn}. Although CNN-based features generally offer lower discriminative power than ViT representations, our decoder brings substantial gains in both anomaly classification and segmentation.
On the MVTec AD dataset, the pixel-level F1-max score improves by 13.7\%, and on VisA, it rises by 21.4\%.
These results demonstrate the ability of our method to refine anomaly maps even with less discriminative CNN features, confirming its backbone-agnostic utility.
For the image-level anomaly classification, AUROC increases from 87.4\% to 90.3\% on MVTec AD and from 56.1\% to 75.2\% on VisA, further underscoring the robustness of our decoder.

\begin{table}[t]
\scriptsize
\centering
\begin{tabular}{lcccc}
\toprule
            \multirow{2}{*}{\textbf{Methods}} &
            \multicolumn{2}{c}{\textbf{MVTec AD}} &
            \multicolumn{2}{c}{\textbf{VisA}} \\
\cmidrule(r){2-3} \cmidrule(l){4-5}
            % \cline{2-5}
            & \textbf{Image-level}& \textbf{Pixel-level} & \textbf{Image-level}& \textbf{Pixel-level} \\
            \midrule
            MuSc (CNN)&  (87.4, 90.3) & (82.1, 36.3) & (56.1, 73.1) & (71.8, 15.0) \\
            % \midrule
            +Ours& (\textbf{90.3}, \textbf{92.1}) & (\textbf{88.7}, \textbf{50.0}) & (\textbf{75.2}, \textbf{78.8}) & (\textbf{80.0}, \textbf{36.4}) \\ 
\bottomrule    
        \end{tabular}
\caption{Ablation study of using CNN features for MuSc in image-level and pixel-level (AUROC, F1-max) on MVTec AD and VisA.}
\label{tab:cnn}
\end{table}

\begin{table}[t]
\resizebox{\linewidth}{!}{
\begin{tabular}{lccccc}
\toprule
            \multirow{2}{*}{\textbf{Methods}} & \multirow{2}{*}{\textbf{Strategy}} &
            \multicolumn{2}{c}{\textbf{MVTec AD}} &
            \multicolumn{2}{c}{\textbf{VisA}} \\
\cmidrule(r){3-4} \cmidrule(l){5-6}
            % \cline{2-5}
            & & \textbf{Image-level}& \textbf{Pixel-level} & \textbf{Image-level}& \textbf{Pixel-level} \\
            \midrule
            \multirow{3}{*}{VCP-CLIP}& ZSAD &(91.7, 95.5) & (49.4, 49.1) & (81.5, 87.3) & (34.7, 31.0) \\
            % \midrule
            & ARD & (91.4, 95.3) & (49.7, 49.2) & (81.4, 87.3) & (34.9, 31.2) \\ 
            & Combined & (\textbf{92.3}, \textbf{95.7}) & (\textbf{52.2}, \textbf{51.6}) & (\textbf{82.0}, \textbf{87.9}) & (\textbf{38.1}, \textbf{32.9}) \\
            \midrule
            \multirow{3}{*}{MuSc*}& ZSAD &(\textbf{97.5}, \textbf{98.7}) & (61.6, 61.2) & (90.1, \textbf{92.8}) & (48.4, 45.0) \\
            % \midrule
            & ARD & (96.9, 98.4) & (61.0, 61.5) & (89.9, 92.2) & (47.9, 44.8) \\ 
            & Combined & (\textbf{97.5}, \textbf{98.7}) & (\textbf{64.4}, \textbf{66.3}) & (\textbf{90.3}, 92.6) & (\textbf{51.3}, \textbf{48.0}) \\
\bottomrule    
        \end{tabular}
        }
\caption{Ablation study of combining the ZSAD method with our AnoRefiner in image-level and pixel-level (F1-max, AP) on MVTec AD and VisA. Best results in bold.}
\label{tab:sum}
\end{table}

\section{Additional ablation study of AnoRefiner.}
\label{app:sum}

To validate the effectiveness of combining the ZSAD method with our AnoRefiner, we compare the original ZSAD output, the output from our ARD alone, and their fused result in \cref{tab:sum}.
The ZSAD method tends to produce high responses even in normal regions, leading to false positives.
In contrast, ARD accurately localizes anomalies but may generate false negatives.
To balance the false positives of ZSAD and the potential false negatives of ARD, we average their anomaly maps.
This operation effectively suppresses both error types and yields superior performance.

\begin{table}
\resizebox{\linewidth}{!}{
\begin{tabular}{lccccc}
\toprule
            \multirow{2}{*}{\textbf{Methods}} & \multirow{2}{*}{\textbf{$u$}} &
            \multicolumn{2}{c}{\textbf{MVTec AD}} &
            \multicolumn{2}{c}{\textbf{VisA}} \\
\cmidrule(r){3-4} \cmidrule(l){5-6}
            % \cline{2-5}
            & & \textbf{Image-level}& \textbf{Pixel-level} & \textbf{Image-level}& \textbf{Pixel-level} \\
            \midrule
            \multirow{3}{*}{VCP-CLIP}& 20 &(91.7, 95.0) & (\textbf{53.0}, \textbf{52.3}) & (81.2, 87.1) & (\textbf{38.4}, \textbf{33.1}) \\
            % \midrule
            & 30 & (\textbf{92.3}, \textbf{95.7}) & (52.2, 51.6) & (\textbf{82.0}, \textbf{87.9}) & (38.1, 32.9) \\
            & 40 & (92.1, 95.4) & (51.9, 51.1) & (81.9, 87.5) & (37.4, 32.6) \\
            \midrule
            \multirow{3}{*}{MuSc}& 20 &(97.3, 98.6) & (\textbf{64.7}, \textbf{66.6}) & (90.0, 92.3) & (\textbf{51.6}, \textbf{48.2}) \\
            % \midrule
            & 30 & (\textbf{97.5}, \textbf{98.7}) & (64.4, 66.3) & (\textbf{90.3}, 92.6) & (51.3, 48.0) \\
            & 40 & (97.4, \textbf{98.7}) & (63.9, 66.0) & (90.2, 92.4) & (50.8, 47.7) \\
\bottomrule    
        \end{tabular}
        }
\caption{Ablation study of different group size $u$ in image-level and pixel-level (F1-max, AP) on MVTec AD and VisA.}
\label{tab:num_u}
\end{table}

\section{Discussion of different group size in PGT.}
\label{app:group}
In Table \ref{tab:num_u}, we evaluate the impact of the group size $u$ in our Progressive Group Test-Time Training (PGT) strategy.
Performance remains consistent across different values of $u$ under both VCP-CLIP and MuSc, with $u=30$ generally offering the best balance between image-level anomaly classification and pixel-level anomaly segmentation.
On the MVTec AD dataset using VCP-CLIP, $u=30$ yields the highest image-level scores (92.3\% F1-max, 95.7\% AP), while $u=20$ achieves the best pixel-level results (53.0\% F1-max, 52.3\% AP).
A similar trend also appears with MuSc on MVTec AD.
On VisA with VCP-CLIP, $u\!=\!30$ again excels in image-level detection (82.0\% F1-max, 87.9\% AP), and $u\!=\!20$ maintains superiority in pixel-level localization (38.4\% F1-max, 33.1\% AP).
Under MuSc, $u\!=\!30$ achieves the highest image-level scores (90.3\% F1-max, 92.6\% AP), while $u\!=\!20$ performs best at the pixel level (51.6\% F1-max, 48.2\% AP).
These results suggest that moderate group sizes ($u$ between 20 and 30) deliver optimal performance, with $u\!=\!30$ favoring image-level classification and $u\!=\!20$ benefiting pixel-level segmentation.
The minimal performance variation within this range confirms the robustness of the group size.

\begin{table}[t]
\resizebox{\linewidth}{!}{
\begin{tabular}{lccccc}
\toprule
            \multirow{2}{*}{\textbf{Methods}} & \multirow{2}{*}{Module} &
            \multicolumn{2}{c}{\textbf{MVTec AD}} &
            \multicolumn{2}{c}{\textbf{VisA}} \\
\cmidrule(r){3-4} \cmidrule(l){5-6}
            & & \textbf{Image-level}& \textbf{Pixel-level} & \textbf{Image-level}& \textbf{Pixel-level} \\
            \midrule
            \multirow{3}{*}{VCP-CLIP}& w/o $\zeta$ &(92.1, 95.5) & (51.9, 51.2) & (81.8, \textbf{87.9}) & (37.7, 32.5) \\
            & w/o $W$ &(91.8, 94.9) & (50.9, 50.2) & (81.4, 87.5) & (37.1, 31.6) \\ 
            & Ours & (\textbf{92.3}, \textbf{95.7}) & (\textbf{52.2}, \textbf{51.6}) & (\textbf{82.0}, \textbf{87.9}) & (\textbf{38.1}, \textbf{32.9}) \\
            \midrule
            \multirow{3}{*}{MuSc}& w/o $\zeta$ &(97.3, 98.4) & (64.0, 65.8) & (90.1, 92.5) & (50.7, 47.4) \\
            & w/o $W$ &(97.0, 98.4) & (62.7, 63.4) & (89.8, 92.0) & (48.9, 46.6) \\
            & Ours & (\textbf{97.5}, \textbf{98.7}) & (\textbf{64.4}, \textbf{66.3}) & (\textbf{90.3}, \textbf{92.6}) & (\textbf{51.3}, \textbf{48.0}) \\
\bottomrule    
        \end{tabular}
        }
\caption{Ablation study of Anomaly-Attention module in image-level and pixel-level (F1-max, AP) on MVTec AD and VisA.}
\label{tab:AAM}
\end{table}

\section{Detail ablation study of Anomaly-Attention module.}
\label{app:AAM}

To evaluate the effectiveness of our proposed Anomaly-Attention (AA) module, we conduct comprehensive ablation studies comparing the complete module against its degraded variants as shown in Table \ref{tab:AAM}. For a fair comparison, all experiments use the same framework configuration, varying only the AA components.

The complete AA module consistently achieves the best performance across both VCP-CLIP and MuSc on the MVTec AD and VisA dataset.
On MVTec AD with VCP-CLIP, it outperforms all ablated variants at both image and pixel levels. Removing the weight component $W$ leads to a noticeable drop in pixel-level metrics, particularly under VCP-CLIP.
The advantage of the full AA module is more evident on the more challenging VisA dataset.
When integrated with MuSc, it sustains strong performance, with a particularly large margin at the pixel level on VisA.
Here, omitting $W$ results in a clear degradation, whereas the complete module maintains robust segmentation quality.
These results above confirm that both the $\zeta$ and $W$ components contribute significantly to the module's effectiveness.
Their joint integration yields the most consistent and accurate performance across different datasets and evaluation settings.

\begin{table}[t]
\resizebox{\linewidth}{!}{
\begin{tabular}{lccccc}
\toprule
            \multirow{2}{*}{\textbf{Methods}} & \multirow{2}{*}{Decoder} &
            \multicolumn{2}{c}{\textbf{MVTec AD}} &
            \multicolumn{2}{c}{\textbf{VisA}} \\
\cmidrule(r){3-4} \cmidrule(l){5-6}
            % \cline{2-5}
            & & \textbf{Image-level}& \textbf{Pixel-level} & \textbf{Image-level}& \textbf{Pixel-level} \\
            \midrule
            \multirow{4}{*}{VCP-CLIP} & DeSTSeg &(90.8, 93.9) & (38.8, 36.1) & (80.8, 86.5) & (26.1,19.2) \\
            % \midrule
            & RealNet &(91.2, 95.1) & (48.0, 47.8) & (81.0, 86.7) & (24.5, 19.4) \\
            & U-Net &(91.1, 95.0) & (48.3, 47.5) & (80.2, 86.4) & (24.1, 17.0) \\
            & ARD (Ours) & (\textbf{91.4}, \textbf{95.3}) & (\textbf{49.7}, \textbf{49.2}) & (\textbf{81.4}, \textbf{87.3}) & (\textbf{34.9}, \textbf{31.2}) \\
            \midrule
            \multirow{4}{*}{MuSc}& DeSTSeg &(96.7, 98.1) & (49.2, 48.4) & (89.8, 92.1) & (26.7, 21.2) \\
            % \midrule
            & RealNet & (96.8, 98.3) & (49.6, 48.9) & (89.8, 91.3) & (29.9, 24.4) \\ 
            & U-Net & (96.4, 98.1) & (49.5, 47.0) & (87.5, 90.8) & (28.6, 22.2) \\ 
            & ARD (Ours) & (\textbf{96.9}, \textbf{98.4}) & (\textbf{61.0}, \textbf{61.5}) & (\textbf{89.9}, \textbf{92.2}) & (\textbf{47.9}, \textbf{44.8})  \\
\bottomrule    
        \end{tabular}
        }
\caption{Comparisons with different decoders in image-level and pixel-level (F1-max, AP) on MVTec AD and VisA.}
\label{tab:decoder}
\end{table}

\begin{table*}[ht]
    \centering
    \setlength{\tabcolsep}{2.2mm}
    \resizebox{1.0\linewidth}{!}{
    \begin{tabular}{llcccccccccccc} 
        \toprule
        \textbf{Datasets} & \textbf{Methods} & \textbf{AUROC-cls} & \textbf{F1-max-cls} & \textbf{AP-cls} & \textbf{AUROC-segm} & \textbf{F1-max-segm} & \textbf{AP-segm} \\ 
        \midrule
        \multirow{4}{*}{MVTec AD~\citeapp{appbergmann2019mvtec}} & AA-CLIP~\citeapp{appma2025aa} & 90.5 & 91.3 & 95.4 & 91.9 & 46.6 & 45.3 \\ 
        & Bayes-PFL~\citeapp{appqu2025bayesian} & 92.3 & 93.1 & 96.7 & 91.8 & - & 48.3 \\
        & RareCLIP~\citeapp{apphe2025rareclip} & 98.2 & 97.6 & \textbf{99.3} & 97.7 & 64.1 & 66.1 \\ 
        & RareCLIP+Ours & \textbf{98.4\footnotesize(+0.2)} & \textbf{97.7\footnotesize(+0.1)} & \textbf{99.3} & \textbf{97.8\footnotesize(+0.1)} & \textbf{65.3\footnotesize(+1.2)} & \textbf{69.0\footnotesize(+2.9)}  \\ 
        \midrule
        \multirow{4}{*}{VisA~\citeapp{appzou2022spot}} & AA-CLIP~\citeapp{appma2025aa} & 84.6 & - & 82.2 & 95.5 & - & - \\
        & Bayes-PFL~\citeapp{appqu2025bayesian} & 87.0 & 84.1 & 89.2 & 95.6 & - & 29.8 \\
        & RareCLIP~\citeapp{apphe2025rareclip} & 94.4 & 90.8 & 95.3 & \textbf{98.8} & 50.9 & 47.5 \\ 
        & RareCLIP+Ours & \textbf{94.7\footnotesize(+0.3)} & \textbf{90.9\footnotesize(+0.1)} & \textbf{95.5\footnotesize(+0.2)} & \textbf{98.8} & \textbf{52.0\footnotesize(+1.1)} & \textbf{49.7\footnotesize(+2.2)}  \\
        \bottomrule
    \end{tabular}
    }
    \caption{Quantitative comparisons with different zero-shot anomaly detection approaches on the MVTec AD and VisA dataset.}
    \label{tab:zsad}
\end{table*}

\begin{table*}[t]
\centering
\footnotesize
\setlength\tabcolsep{0.8mm}
\begin{minipage}{0.49\textwidth}
    \centering
    \textbf{(a) Quantitative results on the MVTec AD dataset in MuSc*.}\\[2pt]
    \resizebox{\textwidth}{!}{%
    \begin{tabular}{lcccccc}
        \toprule
        class & AUROC-cls & F1-max-cls & AP-cls & AUROC-segm & F1-max-segm & AP-segm \\ 
        \midrule
        bottle & 100& 100& 100& 98.9& 79.2& 83.6\\ 
        cable & 99.0& 98.8& 99.5& 96.4& 61.2& 59.0\\ 
        capsule & 95.0& 96.0& 98.8& 98.9& 47.4& 46.8\\ 
        carpet & 99.6& 99.4& 99.9& 99.4& 72.7& 76.3\\ 
        grid & 98.2& 98.4& 99.5& 98.0& 43.0& 35.9\\ 
        hazelnut & 99.5& 97.9& 99.7& 99.4& 72.2& 71.9\\ 
        leather & 100.0& 100.0& 100.0& 99.7& 62.3& 63.2\\ 
        metal\_nut & 100.0& 100.0& 100.0& 90.2& 56.2& 56.0\\ 
        pill & 96.2& 97.4& 99.3& 97.5& 64.8& 66.6\\ 
        screw & 76.7& 88.5& 88.0& 98.5& 34.6& 28.3\\ 
        tile & 100.0& 100.0& 100.0& 97.9& 74.6& 79.4\\ 
        toothbrush & 98.1& 96.6& 99.3& 99.4& 67.0& 62.4\\ 
        transistor & 97.7& 91.0& 97.0& 91.4& 59.6& 58.8\\ 
        wood & 95.9& 96.9& 98.6& 97.3& 68.5& 73.3\\ 
        zipper & 97.6& 97.9& 99.4& 98.4& 62.4& 61.6\\ 
        \midrule
        Mean & 97.1& 97.5 & 98.7 & 97.3& 61.6& 61.2\\ 
        \bottomrule
    \end{tabular}
    }
\end{minipage}
\hfill
\begin{minipage}{0.49\textwidth}
    \centering
    \textbf{(b) Quantitative results on the MVTec AD dataset in MuSc+Ours.}\\[2pt]
    \resizebox{\textwidth}{!}{%
    \begin{tabular}{lcccccc}
        \toprule
        class & AUROC-cls & F1-max-cls & AP-cls & AUROC-segm & F1-max-segm & AP-segm \\ 
        \midrule
        bottle & 100.0& 100.0& 100.0& 98.9& 81.1& 87.5\\ 
        cable & 98.8& 98.4& 99.3& 98.0& 67.0& 69.5\\ 
        capsule & 96.6& 96.4& 99.3& 99.1& 52.1& 49.4\\ 
        carpet & 100.0& 100.0& 100.0& 99.5& 75.6& 82.3\\ 
        grid & 98.4& 96.5& 99.5& 98.1& 45.4& 40.4\\ 
        hazelnut & 99.8& 99.3& 99.9& 99.3& 72.7& 75.7\\ 
        leather & 100.0& 100.0& 100.0& 99.8& 66.4& 70.2\\ 
        metal\_nut & 99.2& 99.0& 99.8& 91.5& 59.0& 61.6\\ 
        pill & 96.5& 96.8& 99.4& 97.7& 66.7& 69.9\\ 
        screw & 76.4& 89.3& 86.9& 98.7& 38.9& 31.8\\ 
        tile & 100.0& 100.0& 100.0& 98.4& 80.1& 85.4\\ 
        toothbrush & 96.5& 96.7& 98.7& 99.5& 67.5& 68.7\\ 
        transistor & 98.3& 92.5& 97.8& 92.9& 62.8& 63.7\\ 
        wood & 97.2& 97.5& 99.0& 97.5& 72.1& 79.1\\ 
        zipper & 99.7& 99.6& 99.9& 98.5& 64.1& 67.9\\ 
        \midrule
        Mean & 97.2& 97.5 & 98.7 & 97.7& 64.4 & 66.3\\
        \bottomrule
    \end{tabular}
    }
\end{minipage}
\caption{Quantitative results on the MVTec AD dataset. All metrics are in $\%$.}
\label{mvtec_detailed_joint}
\end{table*}

\section{Detail analysis of using different decoder.}
\label{app:decoder}
To evaluate the performance of our Anomaly Refinement Decoder (ARD), we compare it against decoders from DeSTSeg \citeapp{appzhang2023destseg}, RealNet \citeapp{appzhang2024realnet} and U-Net \citeapp{appmiccai2015unet} as shown in Table \ref{tab:decoder}.
For a clear evaluation, all refined anomaly maps are used without fusion to the original ZSAD output.
Our ARD consistently outperforms all competing decoders across nearly all metrics on both VCP-CLIP and MuSc methods, with particularly notable gains in pixel-level anomaly segmentation.
On the MVTec AD dataset using VCP-CLIP, ARD achieves 49.7\% F1-max and 49.2\% AP, exceeding U-Net by 1.4\% and 1.7\%, respectively.
With MuSc on the same dataset, ARD attains 61.0\% F1-max and 61.5\% AP, outperforming RealNet by 11.4\% and 12.6\%.
The advantage is more pronounced on the challenging VisA dataset: with VCP-CLIP, ARD reaches 34.9\% F1-max and 31.2\% AP, leading the next best decoder by roughly 10\% and 12\%.
Under MuSc, it achieves 47.9\% F1-max and 44.8\% AP, surpassing RealNet by 18.0\% and 20.4\%.

\section{More comparison with different zero-shot anomaly detection methods}
\label{app:ZSAD}

To evaluate the broad applicability and effectiveness of our proposed AnoRefiner, we integrate it into the current state-of-the-art zero-shot anomaly detection method, RareCLIP~\citeapp{apphe2025rareclip}, and conduct a comprehensive comparative analysis against recent leading methods, AA-CLIP~\citeapp{appma2025aa} and Bayes-PFL~\citeapp{appqu2025bayesian}. The experiments are performed on two standard benchmarks, MVTec AD~\citeapp{appbergmann2019mvtec} and VisA~\citeapp{appzou2022spot}.

Quantitative results in Table \ref{tab:zsad} show that integrating AnoRefiner with RareCLIP (denoted as RareCLIP+Ours) yields consistent improvements across almost all metrics on both datasets.
On the MVTec AD dataset, our refinement yields a notable 2.9\% increase in Average Precision (AP) for anomaly segmentation (from 66.1\% to 69.0\%), indicating more precise segmentation with less false positives and false negatives.
A solid 1.2\% improvement in F1-max-segm (64.1\% to 65.3\%) further confirms better pixel-level segmentation.
Since our AnoRefiner operates chiefly at the pixel level, the image-level metrics have smaller gains (e.g., AUROC-cls from 98.2\% to 98.4\%).
On the more challenging VisA dataset, AnoRefiner achieves a 2.2\% rise in AP-segm (47.5\% to 49.7\%) and a 1.1\% gain in F1-max-segm (50.9\% to 52.0\%).
The consistent upward trend across all metrics underscores the module's robustness in diverse industrial inspection scenarios.

\section{Detailed qualitative results}
\label{app:qualitative}

\begin{figure*}[t]
  \centering
  \includegraphics[width=0.65\linewidth]{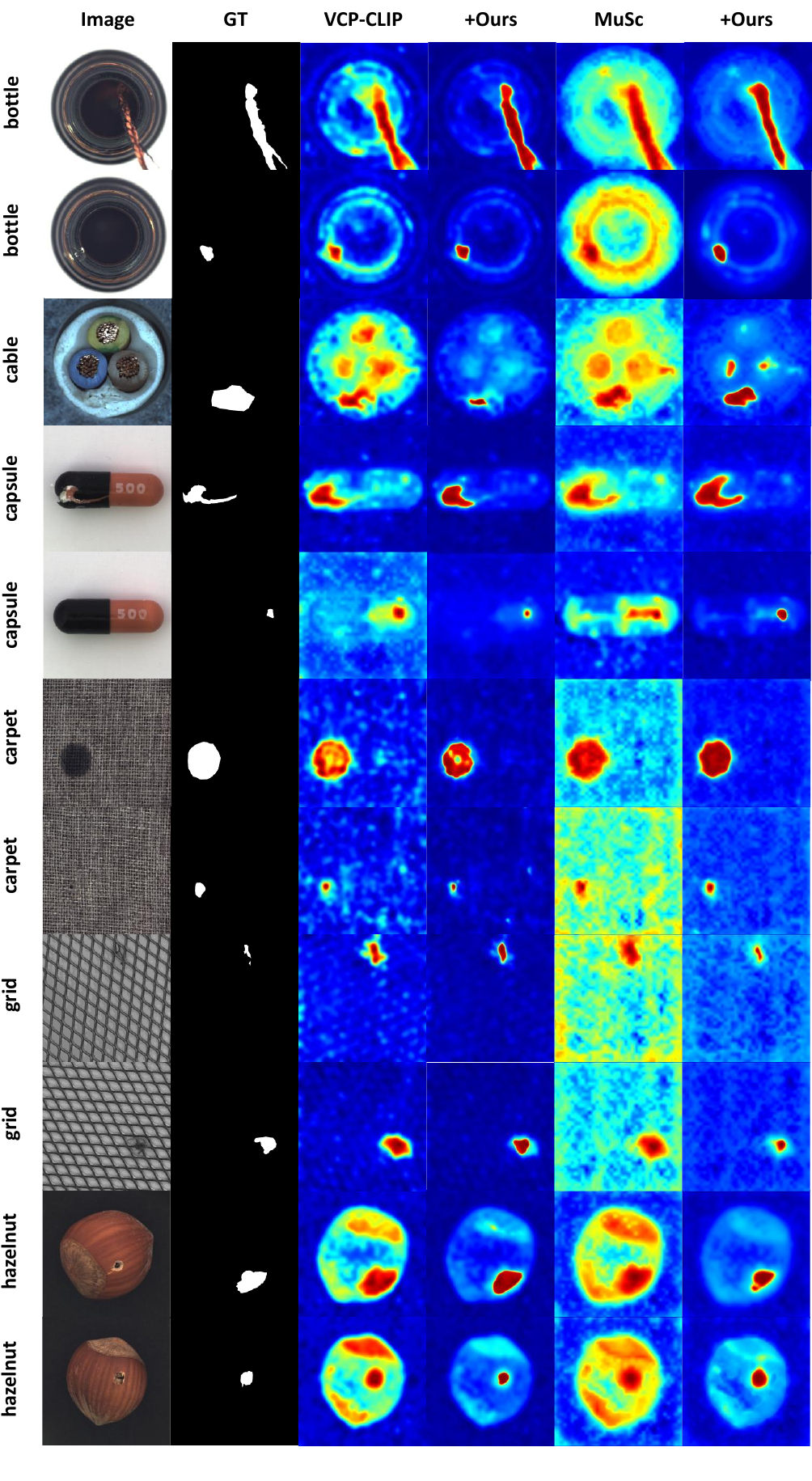}
   \caption{Detailed visualization of anomaly segmentation results on MVTec AD and VisA benchmarks.}
   \label{fig:detail}
\end{figure*}

\begin{figure*}[t]
  \centering
  \includegraphics[width=0.65\linewidth]{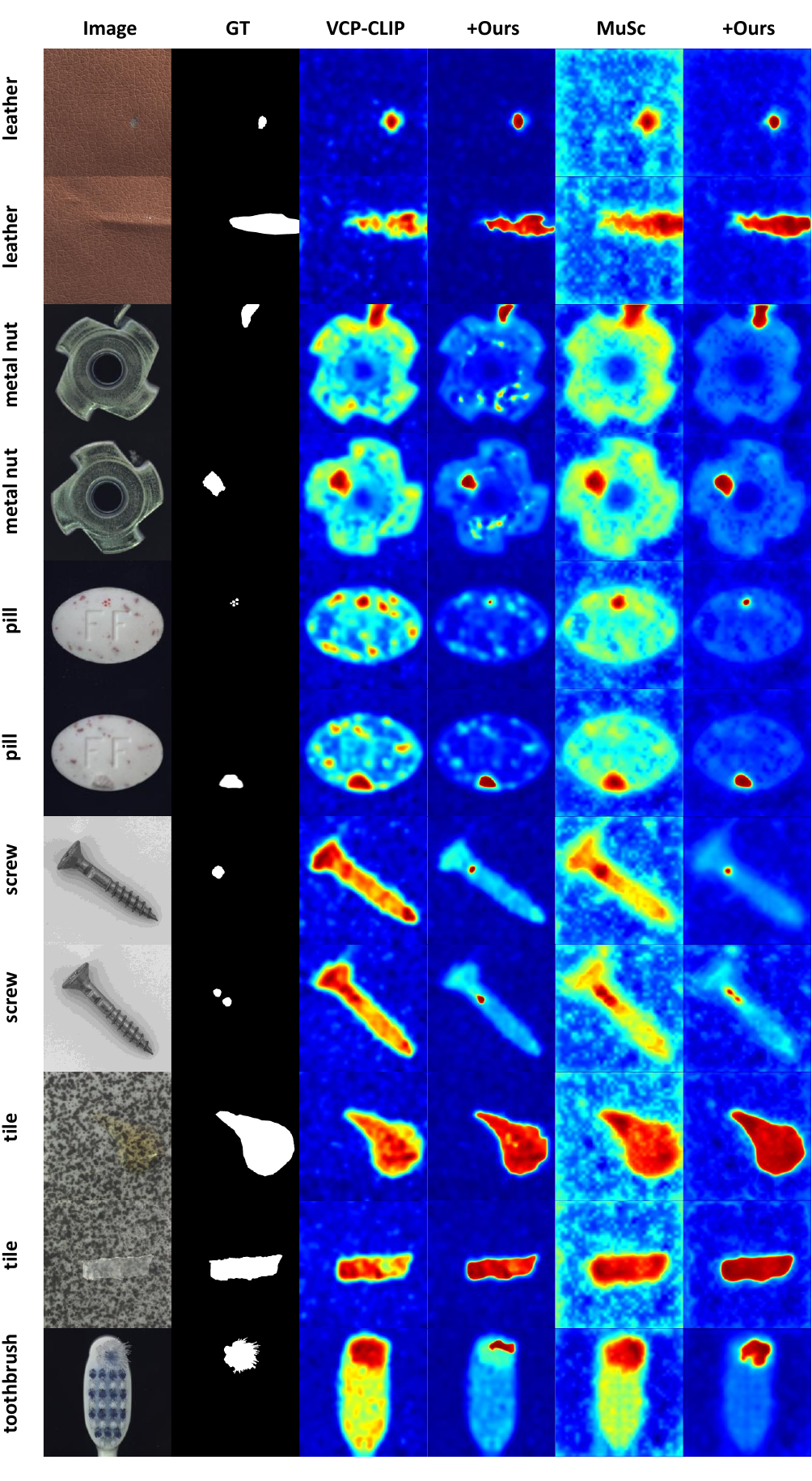}
   \caption{Detailed visualization of anomaly segmentation results on MVTec AD and VisA benchmarks.}
   \label{fig:detail2}
\end{figure*}

\begin{figure*}[t]
  \centering
  \includegraphics[width=0.73\linewidth]{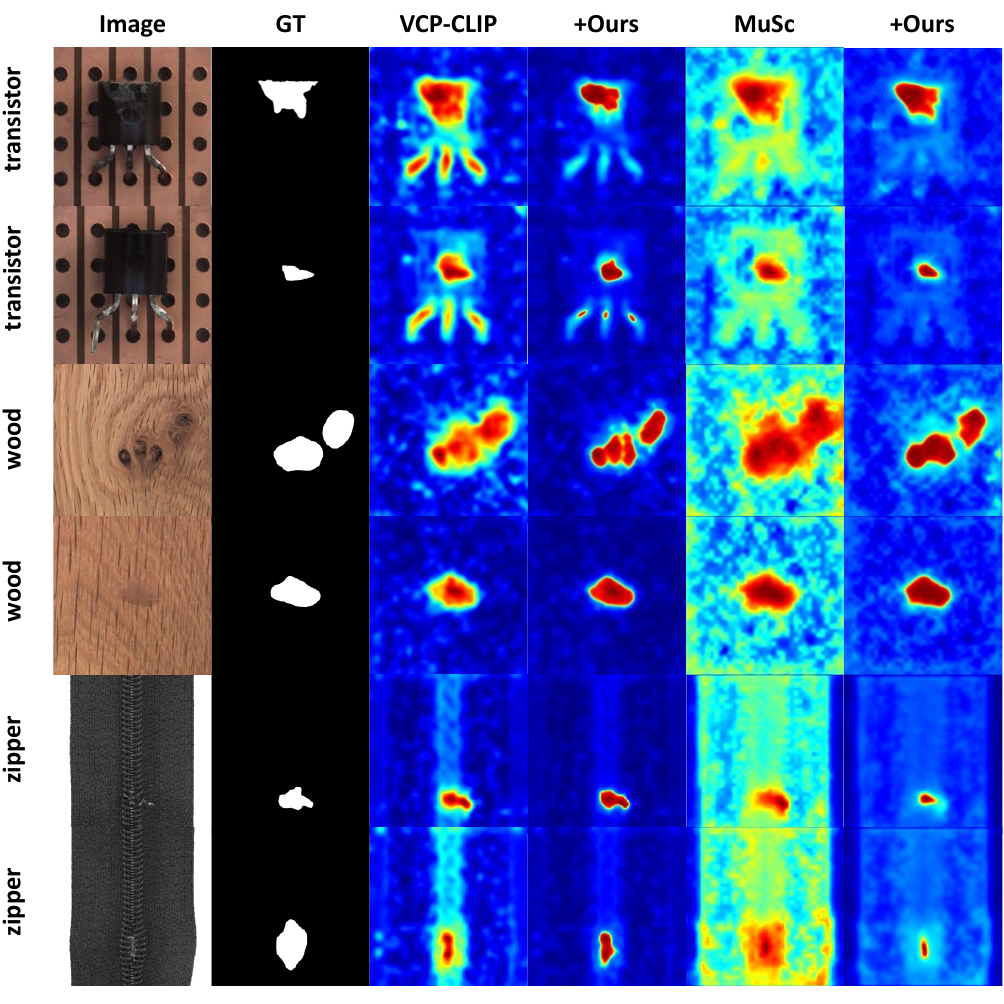}
   \caption{Detailed visualization of anomaly segmentation results on MVTec AD and VisA benchmarks.}
   \label{fig:detail3}
\end{figure*}

From Fig.~\ref{fig:detail} to Fig.~\ref{fig:detail3}, we visualize all categories of the anomaly maps output by MuSc~\citeapp{appli2024musc} and VCP-CLIP~\citeapp{appqu2024vcp} using our AnoRefiner.
Compared to the coarse anomaly maps, our AnoRefiner has fewer false positives.

\section{Detailed quantitative results}
\label{app:quantitative}

We provide a detailed analysis of our AnoRefiner integrated with MuSc on MVTec AD in \cref{mvtec_detailed_joint} (a) and (b). The results demonstrate consistent improvements in pixel-level anomaly segmentation while maintaining strong image-level classification performance. Our approach achieves notable gains in pixel-level metrics: mean AUROC improves by 0.4\%, F1-max increases by 2.8\%, and AP rises by 5.1\%. These improvements are particularly significant in challenging categories with small or complex anomalies—the cable category shows a remarkable 10.5\% relative improvement in AP. Even in categories with high baseline performance like bottle and carpet, our method brings considerable improvements of 3.9\% and 6.0\% in AP, demonstrating broad effectiveness across various anomaly types. Although the screw category shows slight variation in image-level classification metrics, pixel-level segmentation improves by 3.5\% in AP.

Our AnoRefiner consistently enhances pixel-level anomaly segmentation across most MVTec AD categories, with particularly strong gains in challenging scenarios requiring precise localization.

{
    \small
    \bibliographystyleapp{ieeenat_fullname}
    \bibliographyapp{suppl}
}

\end{document}